\documentclass[review]{elsarticle}

\usepackage{lineno,hyperref}
\modulolinenumbers[5]
\usepackage{amsmath}
\usepackage{amssymb}
\usepackage[ruled,vlined]{algorithm2e}
\usepackage{diagbox}
\usepackage{booktabs}
\usepackage{times}
\usepackage{soul}
\usepackage{url}
\usepackage[utf8]{inputenc}
\usepackage[small]{caption}
\usepackage{mathrsfs}
\usepackage{algorithmic}
\usepackage{graphicx}
\usepackage{subfigure}
\usepackage{multirow} 
\usepackage{multicol} 
\usepackage{arydshln}
\usepackage{array}
\usepackage{color}
\newcommand{\RNum}[1]{\uppercase\expandafter{\romannumeral #1\relax}}

\journal{XXX}
\begin{document}
\begin{frontmatter}

\title{Deep Manifold Hashing: A Divide-and-Conquer Approach for Semi-Paired Unsupervised Cross-Modal Retrieval}

\author[mymainaddress]{Yufeng Shi}
\ead{YufengShi17@hust.edu.cn}

\author[myforthaddress,mymainaddress]{Xinge You\corref{mycorrespondingauthor}}
\cortext[mycorrespondingauthor]{Corresponding author}
\ead{youxg@hust.edu.cn}

\author[mysecondaddress]{Jiamiao Xu}
\ead{jiamiaoxu_93@163.com}

\author[mythirdaddress]{Feng Zheng}
\ead{zhengf@sustech.edu.cn}

\author[myforthaddress,mymainaddress]{Qinmu Peng}
\ead{pengqinmu@hust.edu.cn}

\author[myfifthaddress]{Weihua Ou}
\ead{ouweihuahust@gmail.com}


\address[mymainaddress]{School of Electronic Information and Communications, Huazhong University of Science and Technology, Wuhan 430074, China}
\address[myforthaddress]{Research Institute of Huazhong University of Science and Technology in Shenzhen, Shenzhen {\rm 518000}, China}
\address[mysecondaddress]{Department of Deep Learning,
Deeproute Co. Ltd, Shenzhen 518000, China}
\address[mythirdaddress]{Department of Computer Science and Engineering, Southern University of Science and Technology, Shenzhen 518055, China}
\address[myfifthaddress]{School of Big Data and Computer Science, Guizhou Normal University, Guiyang {\rm 550025}, China}

\begin{abstract}
Hashing that projects data into binary codes has shown extraordinary talents in cross-modal retrieval due to its low storage usage and high query speed. Despite their empirical success on some scenarios, existing cross-modal hashing methods usually fail to cross modality gap when fully-paired data with plenty of labeled information is nonexistent. To circumvent this drawback, motivated by the Divide-and-Conquer strategy, we propose Deep Manifold Hashing~(DMH), a novel method of dividing the problem of semi-paired unsupervised cross-modal retrieval into three sub-problems and building one simple yet efficiency model for each sub-problem. Specifically, the first model is constructed for obtaining modality-invariant features by complementing semi-paired data based on manifold learning, whereas the second model and the third model aim to learn hash codes and hash functions respectively. Extensive experiments on three benchmarks demonstrate the superiority of our DMH compared with the state-of-the-art fully-paired and semi-paired unsupervised cross-modal hashing methods.
\end{abstract}


\end{frontmatter}


\section{Introduction}
After decades of rapid development, today's information technology has already ensured people to depict one object in various modalities. Although multi-modal data provides more accessible information to boost the performance of machine learning tasks, it also brings us a challenging yet noteworthy retrieval problem due to the large scale of data and the heterogeneous properties between modalities. Specifically, it is a time-consuming and imprecise job to directly match samples from different modalities, also referred to as cross-modal retrieval~\cite{zhuang2008mining,wang2016comprehensive,peng2017overview}. 

In recent years, cross-modal retrieval has become more valuable than ever due to its ubiquitous applications
in computer version community, including image-text matching~\cite{wu2018joint,wang2020cross}, heterogeneous face image recognition~\cite{chen2014cross,zheng2016hetero} and personalized recommendation~\cite{jiang2017deep,zhu2019r2gan}. Image-text matching is the most common task used for cross-modal retrieval model evaluation, where the database set consists of images and texts fill the query set~(or images query texts). As another typical task, heterogeneous face image recognition retrieves people across different sources such as natural light, near-infrared light and sketches. Obviously, some actions should be taken to remove the modality gap between different modalities before measuring their semantic similarity. 

To tackle cross-modal retrieval problem, some remarkable attempts based on Cross-Modal Hashing~(CMH)~\cite{li2018self,liu2019cyclematch,yao2019online,li2019coupled,2020Joint} have been made, which map data points from different modalities into the same hamming space and use hamming distance as similarity metric among heterogeneous samples. Compared with other Approximate Nearest Neighbor~(ANN) search methods~(e.g., CCL~\cite{peng2017ccl}, ACMR~\cite{wang2017adversarial}) that project data into the same continuous space, hashing based approaches can achieve lower storage usage and higher query speed due to their compact binary codes and specific similarity measurement~\cite{wang2017survey}. The first hashing method in cross-modal retrieval is Cross-modality  Similarity-Sensitive Hashing~(CMSSH)~\cite{bronstein2010data}, which views the foundation of hash functions as a binary classification problem guided by manual annotations. As another representative method, Cross View Hashing~(CVH)~\cite{kumar2011learning} explores the inter-view consistency of fully-paired data to derive effective hash codes. To learn unified hash codes for multimodal objects, both Semantic Topic Multimodal Hashing~(STMH)~\cite{wang2015semantic} and Collective Matrix Factorization Hashing~(CMFH)~\cite{ding2016large} consider common characteristics of fully-paired descriptions in coding procedure. Recently, leveraging the adaptive feature extraction ability of deep learning, Deep Cross-Modal Hashing~(DCMH)~\cite{jiang2017deep} and Self-Supervised Adversarial Hashing~(SSAH)~\cite{li2018self} directly perform hashing code learning from scratch with manual annotations. 

While existing CMH methods have already satisfied the needs of some scenarios, their good performance is usually accompanied by amounts of label annotations~(e.g., category)~\cite{bronstein2010data,jiang2017deep,li2018self} or enough fully-paired training data~(e.g., image-text pairs)~\cite{kumar2011learning,wang2015semantic,ding2016large,li2019coupled}. However, such luxuriant resources not only bring high acquisition cost, but also are scarce in real life. For cross-modal retrieval task, the reality is that objects are not annotated, and only partial objects process fully-paired descriptions. Referring to~\cite{wang2015learning,shen2016semi}, cross-modal retrieval on such data is referred to as semi-paired unsupervised cross-modal retrieval~(SPUCMR). Compared with fully-paired unsupervised cross-modal retrieval, SPUCMR provides more limited information to explore models. To our best knowledge, only five pioneering work are proposed to handle SPUCMR with hashing up to now. Inter-Media Hashing~(IMH)~\cite{song2013inter} and Semi-Paired Hashing~(SPH)~\cite{shen2016semi} directly encode semi-paired and fully-paired samples as hash codes to reveal intra-modal consistency and inter-modal consistency. Later, Partial Multi-Modal Hashing~(P$\text{M}^2$H)~\cite{wang2015learning} and Semi-Paired Discrete Hashing~(SPDH)~\cite{shen2016semi} learn hash codes by ensuring the data consistency via latent subspace learning. Recently, Semi-Paired Asymmetric Deep Cross-Modal Hashing~(SADCH)~\cite{wang2020semi} preserves similarity structure of both the paired points and unpaired points with a cross-view anchor graph. However, as only partial hash codes in their methods are encoded with common characteristics of fully-paired descriptions, their hamming space is not modality-invariant, which thus leads to the residue of modality gap.  

To this end, we propose a novel semi-paired unsupervised cross-modal method, named Deep Manifold Hashing~(DMH), with the Divide-and-Conquer strategy~[~\cite{cormen2009introduction}, chapter 4]. Instead of constructing a convoluted ``unified'' model, we split the problem of semi-paired unsupervised cross-modal retrieval into three sub-problems and devise one model for each sub-problem. Specifically, the first model is constructed for complementing semi-paired data to extract modality-invariant features of fully-paired descriptions based on manifold learning, whereas the second model aims to encode features as binary codes by optimizing KL-divergence. Finally, the third model is designed to learn hash functions for different modalities using deep neural networks. 

To summarize, our main contributions are threefold:
\begin{itemize}
\item Motivated by the Divide-and-Conquer strategy, a novel method named DMH, is proposed for semi-paired unsupervised cross-modal retrieval~(SPUCMR). Three sub-problems in connection with SPUCMR are pinpointed and three models are constructed to solve each of them.
\item In practice, un-paired data is usually just discarded for convenience, which leads researchers to handle unsupervised cross-modal retrieval~(UCMR). Fortunately, our DMH can manage UCMR and SPUCMR simultaneously with small changes.
\item Extensive experiments on MIRFLICKR-25K, MS COCO and NUS WIDE datasets demonstrate that our DMH can remove modality gap more effectively than other methods, thus boosting the retrieval performance.
\end{itemize}

The rest of this paper is organized as follows. Sect.~2 introduces related studies on cross-modal hashing and manifold learning. Sect.~3 presents the proposed DMH method and its optimization in detail. The experimental results and analyses are reported in Sect.~4. Finally, Sect.~5 concludes this paper.

\section{Related Work}
In this section, cross-modal hashing methods are briefly reviewed. To make readers easier understand our work, some knowledge on manifold learning is also introduced.
\subsection{Cross-modal hashing}
Cross-modal hashing has made remarkable progress in handling the problem of cross-modal retrieval, and this type of methods can be roughly divided into two major categories supervised approaches~\cite{zhang2014large,lin2015semantics,wang2016multimodal,deng2018triplet,yu2019discriminative} and unsupervised approaches~\cite{zhou2014latent,wang2015semantic,ding2016large,liu2017cross,li2019coupled}.

Supervised cross-modal hashing methods fully utilize the semantic labels in the process of learning binary codes to reduce the modality gap, and can usually achieve higher retrieval accuracy than unsupervised ones. For example, semantic labels can guide hash codes to preserve semantic mutual similarity such as pair-wise~\cite{zhang2014large,yang2017pairwise}, triplet-wise~\cite{deng2018triplet,liu2019ranking} or multi-wise similarity relations~\cite{lin2015semantics,shi2019equally}. For pairwise similarity, Semantic Correlation Maximization~(SCM)~\cite{zhang2014large} constructs the pairwise semantic similarity by the cosine similarity between semantic labels. For triplet-wise, Triplet-based Deep Hashing~(TDH)~\cite{deng2018triplet} utilizes the triplet labels, which describe the relative relationships among three samples as supervision to capture more semantic correlations. To cover multi-wise similarity, Semantics-Preserving Hashing~(SePH)~\cite{lin2015semantics} minimizes the KL-divergence between distributions of semantic labels and hash codes. And in the meantime, semantic labels also provide category properties of samples, which is the criteria of aggregating intra-class data points. Therefore, Multimodal Discriminative Binary Embedding~(MDBE)~\cite{wang2016multimodal} attempts to learn hash functions based on classification to reveal the correspondence between hash codes and labels. As another representative method, Discriminative Supervised Hashing~(DSH)~\cite{yu2019discriminative} also regards hash codes as easily classified features. 

For the second category, unsupervised cross-modal hashing methods obtain hash codes and train hash functions without resorting to manual semantic labels. While relying on data distributions as supervisor~\cite{wang2015semantic,su2019deep}, most of existing unsupervised cross-modal hashing methods also exploit the common characteristics of fully-paired data to cross modality gap~\cite{liu2017cross,li2019coupled}. For example, Latent Semantic Sparse Hashing~(LSSH)~\cite{zhou2014latent} aims to capture high-level latent semantic information and generate unified codes for image-text pairs. CMFH~\cite{ding2016large} and STMH~\cite{wang2015semantic} both attempt to learn unified hash codes for different modal descriptions of one multimodal object in the shared latent semantic space. Another commonly used approach, Fusion Similarity Hashing~(FSH)~\cite{liu2017cross} preserves fusion similarity among multimodal objects.  

To fit the reality that objects are not annotated and only partial objects process fully-paired descriptions in real applications, few precursory hashing methods~\cite{song2013inter,wang2015learning,shen2016semi,wang2020semi} are proposed to conduct SPUCMR. As the earliest work, IMH~\cite{song2013inter} encodes images and their relevant texts as similar hash codes, while preserves local structural information to obtain embeddings for un-paired samples. As another early attempt, P$\text{M}^2$H~\cite{wang2015learning} learns hash codes for images and texts respectively via latent subspace learning and graph Laplacian. According to intra-modal consistency and inter-modal consistency, SPH~\cite{shen2016sem} and SPDH~\cite{shen2016semi} explore the underlying structure of the constructed common latent subspace, where both paired and unpaired samples are well aligned. Recently, SADCH~\cite{wang2020semi} constructs a novel cross-view graph to encode image-text pairs, un-paired images and un-paired texts separately. Although they pinpointed accurately the data consistency, only partial hash codes in their methods are encoded with common characteristics of fully-paired descriptions, leading to the residue of modality gap. In addition, almost all these existing SPUCMR methods are based on hand-crafted features, which can not be self-adapting in optimization.

Nowadays, deep learning has become effectual to adaptively extract features from scratch for impelling cross-modal retrieval performance. As a beginning of deep cross-modal hashing, Deep Visual-Semantic Hashing~(DVSH)~\cite{cao2016deep} combines AlexNet~\cite{krizhevsky2012imagenet} over images, Recurrent Neural Networks~(RNN) over sentences to learn similarity-preserving hash codes. To bridge the modality gap, SSAH~\cite{li2018self} leverages two adversarial networks to maximize the semantic correlation and consistency of the representations between different modalities. To fit real world settings, unsupervised deep hashing methods aim to learn the modality correlation depending on correspondences of fully-paired data. Unsupervised Deep Cross Modal Hashing~(UDCMH)~\cite{wu2018unsupervised} integrates deep learning and matrix factorization with binary latent factor models to generate unified binary codes for fully-paired data. Later, Unsupervised Cycle Hashing~(UCH)~\cite{li2019coupled} devises pair-coupled generative adversarial networks to build two cycle networks in an unified framework to adequately explore paired correspondences between modalities. However, such fully-paired data is still infrequent in daily life.
\subsection{Manifold learning}
Manifold learning constructs a low-dimensional manifold through precisely describing and preserving local geometric information of the high-dimensional space~\cite{tenenbaum2000global,roweis2000nonlinear,belkin2006manifold}. For example, Locally Linear Embedding~(LLE)~\cite{roweis2000nonlinear} assigns each data point and its neighbors to lie on or close to a locally linear patch of the manifold. Recently, since one object can be easily depicted in various views or modalities, multi-view manifold learning methods~\cite{zong2017multi,zhao2018multi,xiao2019multi,feng2020multi} are proposed to find a common manifold that can reveal information of input views. Multi-View Manifold Learning with Locality Alignment~(MVML-LA)~\cite{zhao2018multi} finds a common latent space and aligns neighbors in all original views to be geometrically close to each other. Multiview Locality Low-rank Embedding~(MvL$^2$E)~\cite{feng2020multi} fully utilizes correlations between multi-views by adopting low-rank representations to capture a common low-dimensional embedding among views.

Benefiting from manifold learning, researchers can explore the inherent structure of multi-modal data and further boost the performance of cross-modal retrieval. For example, Parallel Field Alignment Retrieval~(PFAR)~\cite{mao2013parallel} considers cross-modal retrieval as a manifold alignment problem using parallel fields from the perspective of vector fields. A metric learning framework named Multi-ordered Discriminative Structured Subspace Learning~(MDSSL)~\cite{zhang2016cross} considers the multi-order statistical features which lie on the different Euclidean spaces and Riemannian manifolds. To accelerate retrieval and save storage space, methods that combine manifold learning and hashing emerge including Hetero-Manifold Regularisation~(HMR)~\cite{zheng2016hetero} and Supervised Discrete Manifold-Embedded Cross-Modal Hashing~(SDMCH)~\cite{luo2018sdmch}. HMR integrates multiple uni-modal and cross-modal sub-manifolds into a common manifold and introduces hetero-manifold regularised hash function learning. SDMCH not only exploits the nonlinear manifold structure of data and correlations among heterogeneous multiple modalities, but also fully utilizes the manual annotations. However, there still remains a need for a cross-modal hashing method that can employ manifold learning and hashing simultaneously in semi-paired unsupervised cross-modal retrieval scenario.

\section{Deep Manifold Hashing and its optimization}
In this section, we first present the problem definition, and then introduce the motivation and general idea of our proposed DMH in detail. The optimization procedure is finally given. For illustration purposes, we apply our method in two most frequently-used modalities image and text.
\subsection{Notation and problem definition}
Matrix and vector used in this paper are represented by boldface uppercase letter~(e.g., $\boldsymbol{Y}$) and boldface lowercase letter~(e.g., $\boldsymbol{y}$), respectively. $\left\|\cdot\right\|$ denotes the 2-norm of vectors. $sign(\cdot)$ is defined as sign function, which outputs 1 if its input is positive else outputs -1. Let $\boldsymbol{X}^1=\left[ \boldsymbol{x}_{1}^{1},\boldsymbol{x}_{2}^{1},...,\boldsymbol{x}_{n_m}^{1},\boldsymbol{x}_{n_m+1}^{1},\boldsymbol{x}_{n_m+2}^{1},...,\boldsymbol{x}_{n_m+n_1}^{1} \right]$ and $\boldsymbol{X}^2=\left[ \boldsymbol{x}_{1}^{2},\boldsymbol{x}_{2}^{2},...,\boldsymbol{x}_{n_m}^{2},\boldsymbol{x}_{n_m+1}^{2},\boldsymbol{x}_{n_m+2}^{2},...,\boldsymbol{x}_{n_m+n_2}^{2} \right]$ symbolize images and texts of training set, where $\boldsymbol{x}_{i}^{1}\in \mathbb{R}^{d_1}$, $\boldsymbol{x}_{j}^{2}\in \mathbb{R}^{d_2}$, $n_m$ is the number of image-text pairs, $n_1$ and $n_2$ are the numbers of un-paired data in images and texts, respectively. Hence, the number of objects in training set is $n$, where $n=n_m+n_1+n_2$.

Given bit-length $c$ and $n$ objects for training, the goal of hashing methods is to learn hash functions $g^{1}\!\left(\!\boldsymbol{x}_{i}^{1}\!\right)\!:\mathbb{R}^{d_{1}}\!\rightarrow\!\left\{\! -\!\text{1,}~1\!\right\}^c$ and $g^{2}\!\left(\!\boldsymbol{x}_{j}^{2} \!\right)\!:\mathbb{R}^{d_{2}}\!\rightarrow\!\left\{\!-\!\text{1,}~1\!\right\}^c$, which map images and texts as hash codes $\boldsymbol{h}_{i}^{1}$ and $\boldsymbol{h}_{j}^{2}$ in the same Hamming space respectively. Similar to most existing hashing methods~\cite{jiang2017deep,cao2018cross,li2018self}, we first construct approximate hash functions $f^{1}\left( \theta _{1};\boldsymbol{x}_{i}^{1} \right)$ and $f^{2}\left( \theta _{2};\boldsymbol{x}_{j}^{2} \right)$, and use $sign(\cdot)$ to build our hash functions:
\begin{align}
    g^1\left( \boldsymbol{x}_{i}^{1} \right)&=sign\left( f^1\left( \theta _1;\boldsymbol{x}_{i}^{1} \right) \right)\nonumber\\
    g^2\left( \boldsymbol{x}_{j}^{2} \right)&=sign\left( f^2\left( \theta _2;\boldsymbol{x}_{j}^{2} \right) \right).
\end{align}

In this paper, we adopt $D(\boldsymbol{h}_{i}^{1},\boldsymbol{h}_{j}^{2})=\frac{1}{4}\left\| \boldsymbol{h}_{i}^{1}-\boldsymbol{h}_{j}^{2} \right\| ^{2}$ to calculate Hamming distance between hash codes $\boldsymbol{h}_{i}^{1}$ and $\boldsymbol{h}_{j}^{2}$. Meanwhile, Hamming distance $D( \boldsymbol{h}_{i}^{1},\boldsymbol{h}_{j}^{2})$ needs to indicate the semantic similarity between $\boldsymbol{x}_{i}^{1}$ and $\boldsymbol{x}_{j}^{2}$.

\subsection{Deep Manifold Hashing}
Un-annotated semi-paired data emerge everywhere, posing a new challenge of handling modality-invariant features, hash codes and hashing functions simultaneously~\cite{li2019coupled,cao2018cross,wang2020semi}. Inspired by the Divide-and-Conquer strategy, the problem of semi-paired unsupervised cross-modal retrieval can be split into three sub-problems: (\uppercase\expandafter{\romannumeral1}) extracting modality-invariant features, (\uppercase\expandafter{\romannumeral2}) encoding hash codes and (\uppercase\expandafter{\romannumeral3}) fitting hash functions.

\begin{figure*}[!htp]
\centering
\includegraphics[scale=0.42]{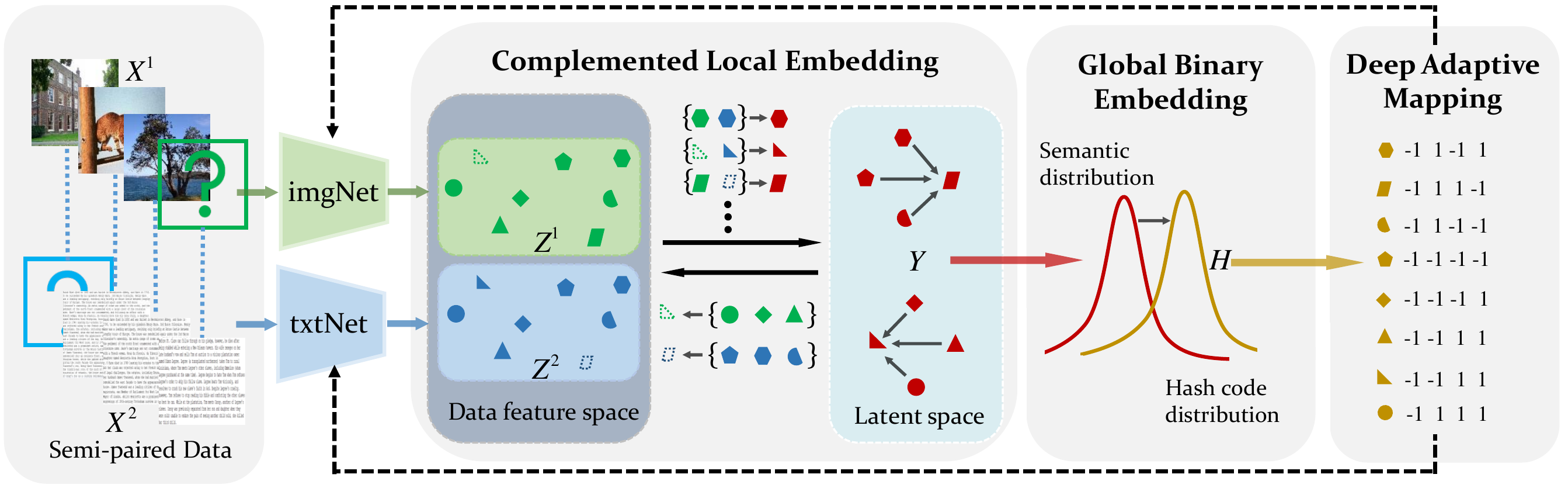}
\caption{An overview of Deep Manifold Hashing~(DMH). Different markers denote different classes and different colors indicate different spaces. The hollow shapes in $\boldsymbol{Z}^v|_{v=1}^{2}$ represent descriptions complemented by Complemented Local Embedding, while the solid shapes in modality-invariant space $\boldsymbol{Y}$ denote the fully-paired level objects. Following this, Global Binary Embedding transforms modality-invariant features into hash codes with the consideration of global similarity. Finally, Deep Adaptive Mapping builds hash functions to fit the correspondence between data of different modalities and unified hash codes.}
\label{fig1}
\end{figure*}

In this paper, three models, Complemented Local Embedding~(CLE), Global Binary Embedding~(GBE) and Deep Adaptive Mapping~(DAM), are developed respectively to solve above three sub-problems, and we finally integrate three models in a joint manner to handle the problem of SPUCMR. The combined model is termed as DMH, and its schematic illustration is shown in Figure~\ref{fig1}.

\subsubsection{Complemented Local Embedding}
Descriptions of the same objects~(i.e., fully-paired descriptions) share common characteristics~(i.e., modality-invariant features) that are the key to remove modality gap. However, data are not always collected in fully-paired form, thus impeding the extraction of common characteristics. Motivated by manifold learning~\cite{roweis2000nonlinear,chang2006robust}, we employ the correspondence between modality-invariant space and original description spaces to complement semi-paired descriptions as fully-paired descriptions, and extract common characteristics of complemented fully-paired descriptions and existing fully-paired descriptions with local alignment constraint. 

Specifically, we assume that image-text pairs are descriptions of objects in the
modality-invariant space. To reduce computational complexity, we use the features $\boldsymbol{z}_{i}^{v}$ extracted by imgNet and txtNet to represent descriptions $\boldsymbol{x}_{i}^{v}$. Hence, we reconstruct $n_m$ image-text pairs~(i.e., $n_m$ objects with both image and text descriptions) based on the $d$-dimensional modality-invariant features $\boldsymbol{Y}\!=\!\left\{ \boldsymbol{y}_i \right\} _{i=1}^{n_m}$ via minimizing the $L_2$ distance between $\boldsymbol{z}_{i}^{v}$ and $\boldsymbol{Q}^v\boldsymbol{y}_i$:
\begin{small}
\begin{align}
\ell _1&\!=\!\underset{\boldsymbol{Q}^v,\boldsymbol{Y}}{arg\min}\!\frac{1}{2n_m}\!\sum_{i=1}^{n_m}\!{\sum_{v=1}^2\!{\lVert \boldsymbol{z}_{i}^{v}\!-\!\boldsymbol{Q}^v\boldsymbol{y}_i \rVert ^2}},
\end{align}
\end{small}
where the projection functions $\boldsymbol{Q}^v|_{v=1}^{2}$ connect the description spaces of different modalities and the modality-invariant space. 

To establish the correspondence between  modality-invariant space and original description spaces, the modality-invariant space should reveal local data consistency. Consequently, $\boldsymbol{Q}^v|_{v=1}^{2}$ should also guide modality-invariant features to reflect the local geometric characteristics of description spaces. Therefore, $\boldsymbol{y}_i$ can be viewed as the linear combination of $k$ nearest neighbors in the manifold. To this end, Eq.~(2) can be formulated as:
\begin{small}
\begin{align}
\ell _1&=\underset{\boldsymbol{Q}^v,\boldsymbol{Y},\boldsymbol{W}}{arg\min}\frac{1}{2n_m}\sum_{i=1}^{n_m}{\sum_{v=1}^2{\lVert \boldsymbol{z}_{i}^{v}-\boldsymbol{Q}^v\boldsymbol{y}_i \rVert ^2}}\\\nonumber 
&+\frac{\lambda}{n_m}\sum_{i=1}^{n_m}{\lVert \boldsymbol{y}_i-\sum_{j=1}^k{\boldsymbol{W}_{ij}\boldsymbol{y}_{ij}} \rVert ^2}\\\nonumber 
&s.t.\sum_{j=1}^k{\boldsymbol{W}_{ij}}=1,
\end{align}
\end{small}
where $\boldsymbol{W}_{ij}$ measures the proportion of $k$ nearest neighbors of $\boldsymbol{y}_i$, and $\lambda$ controls the contribution of local data consistency. For image-text pairs, the $k$ nearest neighbors of $\boldsymbol{y}_i$ are selected from the intersection of nearest neighbors in both description spaces.

In order to strengthen the generalization ability, a regularization term ${\lVert \boldsymbol{y}_i \rVert ^2}$ is introduced. Moreover, we also constrain mapping functions $\boldsymbol{Q}^v|_{v=1}^{2}$ to be orthogonal for non-trivial solutions as follows:
\begin{small}
\begin{align}
\ell _1&=\underset{\boldsymbol{Q}^v,\boldsymbol{Y},\boldsymbol{W}}{arg\min}\frac{1}{2n_m}\sum_{i=1}^{n_m}{\sum_{v=1}^2{\lVert \boldsymbol{z}_{i}^{v}-\boldsymbol{Q}^v\boldsymbol{y}_i \rVert ^2}}\\\nonumber &+\frac{\lambda}{n_m}\sum_{i=1}^{n_m}{\lVert \boldsymbol{y}_i-\sum_{j=1}^k{\boldsymbol{W}_{ij}\boldsymbol{y}_{ij}} \rVert ^2}\\\nonumber
&+\eta\!\sum_{i=1}^{n_m}\!{\lVert \boldsymbol{y}_i \rVert ^2}\\\nonumber
&s.t.\sum_{j=1}^k{\boldsymbol{W}_{ij}}=1,\left(\!\boldsymbol{Q}^v\!\right)^T\!\boldsymbol{Q}^v\!=\!\boldsymbol{I},
\end{align}
\end{small}
where $\eta$ is a non-negative trade-off parameter weighting the importance of regularization term, and $\boldsymbol{I}$ indicates the identity matrix.

The modality-invariant features of image-text pairs can be extracted by optimizing $\ell _1$, which acts as the cornerstone in fully-paired scenario. Nevertheless, not all objects could own descriptions in both modalities. Fortunately, manifold learning also provides a feasible solution to tackle incomplete-modality problem via the correspondence between origin high-dimensional description spaces and latent low-dimensional modality-invariant manifold. Therefore, we complement descriptions of incomplete objects~(i.e., $n_1$ objects with only image descriptions and $n_2$ objects with only text descriptions) according to the local linearity discovered on latent manifold:
\begin{small}
\begin{align}
\boldsymbol{\bar{z}}_{i}^{v}=\left\{ \begin{matrix}	\boldsymbol{z}_{i}^{v}&		\text{if}\,\,\boldsymbol{x}_{i}^{v}\,\,\text{exists};\\	\sum_{j=1}^k{\boldsymbol{W}_{ij}\boldsymbol{z}_{j}^{v}}&		\text{otherwise},\\\end{matrix} \right.
\end{align}
\end{small}
where $\boldsymbol{z}_{j}^{v}$ corresponds to the descriptions of fully-paired object $\boldsymbol{y}_{j}$ in modality $v$ to guarantee its existence. And the $k$ nearest neighbors of un-paired data are also selected from these fully-paired data based on intra-modality consistency. 

Finally, the objective of Complemented Local Embedding~(CLE) is represented as:
\begin{small}
\begin{align}
\ell _1&=\underset{\boldsymbol{Q}^v,\boldsymbol{Y},\boldsymbol{W},\boldsymbol{\bar{Z}}}{arg\min}\frac{1}{2n}\sum_{i=1}^{n}{\sum_{v=1}^2{\lVert \boldsymbol{\bar{z}}_{i}^{v}-\boldsymbol{Q}^v\boldsymbol{y}_i \rVert ^2}}\\\nonumber 
&+\frac{\lambda}{n}\sum_{i=1}^{n}{\lVert \boldsymbol{y}_i-\sum_{j=1}^k{\boldsymbol{W}_{ij}\boldsymbol{y}_{ij}} \rVert ^2}\\\nonumber
&+\eta\sum_{i=1}^{n}{\lVert \boldsymbol{y}_i \rVert ^2}\\\nonumber
&s.t.\left( \boldsymbol{Q}^v \right) ^T\boldsymbol{Q}^v=\boldsymbol{I},\sum_{j=1}^k{\boldsymbol{W}_{ij}}=1,
\end{align}
\end{small}
where $n$ indicates the sum of $n_m$ image-text pairs, $n_1$ semi-paired images and $n_2$ semi-paired texts. Meanwhile, the dimension of modality-invariant features $d$ and the number of nearest neighbors $k$ are set to 512 and 3 by default.

\subsubsection{Global Binary Embedding}
The modality-invariant feature learned by CLE model lies on a continuous manifold, which 
can not meet the requirement of binarization and $c$ bit length. To obtain discrete hash code with global data consistency, inspired by~\cite{lin2015semantics,ma2018global}, Global Binary Embedding~(GBE) attempts to estimate the distribution of correlations among data points in Hamming space $\boldsymbol{S}^H$ by minimizing its distance with the distribution of correlations among data points in modality-invariant space $\boldsymbol{S}^Y$. 

Specifically, we define $\boldsymbol{S}^Y_{i,j}$ as the $i$-th row $j$-th column element of $\boldsymbol{S}^Y$ and it measures the similarity between $\boldsymbol{y}_i$ and $\boldsymbol{y}_j$ in the modality-invariant space. And we compute inner products of modality-invariant features and utilize them to measure $\boldsymbol{S}^Y_{i,j}$:
\begin{small}
\begin{align}
\boldsymbol{S}^Y_{i,j}=\frac{\boldsymbol{y}_{i}^{T} \boldsymbol{y}_{j}}{\varSigma _{i=1}^{n}\varSigma _{j=\text{1,}j\ne i}^{n}\boldsymbol{y}_{i}^{T} \boldsymbol{y}_{j}}\\ \nonumber
s.t.\varSigma _{i=1}^{n}\varSigma _{j=\text{1,}j\ne i}^{n}\boldsymbol{S}^Y_{i,j}=1.
\end{align}
\end{small}
For $\boldsymbol{S}^H$, we define $\boldsymbol{S}^H_{i,j}$ as the $i$-th row $j$-th column element of $\boldsymbol{S}^H$ and it measures the similarity between hash codes $\boldsymbol{h}_i$ and $\boldsymbol{h}_j$ in Hamming space. Therefore, following~\cite{maaten2008visualizing,zhang2019deep}, we assume their Hamming distances obey a Student T-distribution with one degree of freedom, and thus utilize it to measure $\boldsymbol{S}^H_{i,j}$:
\begin{small}
\begin{align}
\boldsymbol{S}^H_{i,j}=\frac{\left( 1+D\left( \boldsymbol{h}_i,\boldsymbol{h}_j \right) \right) ^{-1}}{\sum_{i=1}^n{\sum_{j=\text{1,}j\ne i}^n{\left( 1+D\left( \boldsymbol{h}_i,\boldsymbol{h}_j \right) \right) ^{-1}}}}.
\end{align}
\end{small}
To minimize the distance between $\boldsymbol{S}^Y$ and $\boldsymbol{S}^H$, we employ the Kullback-Leibler divergence as metric, whose cost is formulated as follows:
\begin{small}
\begin{align}
\ell _{21}&=\underset{\boldsymbol{H}}{arg\min}KL\left( \boldsymbol{S}^Y \parallel \boldsymbol{S}^H \right)\\\nonumber
&=\sum\nolimits_{i=1}^{n}{\sum\nolimits_{j=1,j\ne i}^{n}{\boldsymbol{S}^Y_{i,j}\log \left( \frac{\boldsymbol{S}^Y_{i,j}}{\boldsymbol{S}^H_{i,j}} \right)}}.
\end{align}
\end{small}
Note that, Eq.~(9) is non-convex due to the discrete attribute of $\boldsymbol{H}$, which blocks its direct optimization. Therefore, the binary $\boldsymbol{H}$ is relaxed as continuous $\boldsymbol{\hat{H}}$ in Eq.~(8), and quantization loss is also introduced to diminish its impact:
\begin{small}
\begin{align}
\ell _{22}&=\underset{\boldsymbol{\hat{H}}}{arg\min}\lVert \hat{\boldsymbol{h}_i}-\boldsymbol{h}_i \rVert ^{2},
\end{align}
\end{small}
where $\boldsymbol{h}_i$ indicates the result of $sign(\hat{\boldsymbol{h}_i})$.

Finally, the objective function of GBE is represented as:
\begin{small}
\begin{align}
\ell _{2}&= \ell _{21} + \gamma\ell _{22}\\\nonumber
&=\underset{\hat{\boldsymbol{H}}}{arg\min}\sum\nolimits_{i=1}^{n}{\sum\nolimits_{j=1,j\ne i}^{n}{\boldsymbol{S}^Y_{i,j}\log \left( \frac{\boldsymbol{S}^Y_{i,j}}{\hat{\boldsymbol{S}^H_{i,j}}} \right)}}\\\nonumber
&+\gamma\lVert \hat{\boldsymbol{h}_i}- \boldsymbol{h}_i \rVert ^{2}.
\end{align}
\end{small}
where $\hat{\boldsymbol{S}^H_{i,j}}$ indicates the $\boldsymbol{S}^H_{i,j}$ calculated with $\hat{\boldsymbol{h}}$, and $\gamma$ is a trade-off parameter to balance the effect of the two terms. When the optimization procedure of GBE is finished, we use $sign(\cdot)$ to handle $\hat{\boldsymbol{H}}$ and obtain hash codes $\boldsymbol{H}$.

\subsubsection{Deep Adaptive Mapping}
Once global semantics-preserving hash codes $\boldsymbol{H}$ are obtained, the only item in to-do list is to learn modality-specific hash functions, which aim to produce hash codes for new samples. To build the relationship between hash codes and data of different modalities, we construct Deep Adaptive Mapping with two deep neural networks~(i.e., imgNet and txtNet), making it possible to fit complex mapping relations and  adaptively extract features as inputs of CLE model.

Specifically, we modify CNN-F~\cite{chatfield2014return} to build imgNet. To obtain $c$ bit length hash codes, the last fully-connected layer in origin CNN-F is changed to a $c$-node fully-connected layer. For text, we first use the multi-scale network in ~\cite{li2018self} to extract multi-scale information and a two-layer MLP whose nodes are 4096 and $c$ to transform them into hash codes. Except the activation function of last layers is $tanh(\cdot)$ to approximate $sign(\cdot)$ function, other layers use ReLU as activation functions. To improve generalization performance, Local Response Normalization~(LRN)~\cite{krizhevsky2012imagenet} is applied between layers of all MLPs. It should be noted that the backbones of imgNet and txtNet are employed only for illustrative purposes, and any other basic networks can be easily adopted by our DMH.

To fit hash functions, the objective function of Deep Adaptive Mapping~(DAM) is as follows:
\begin{small}
\begin{align}
\ell _3=\underset{\theta _1,\theta _2}{\min}\sum_{i=1}^{n_m+n_1}{\lVert f^1\left( \theta _1;\boldsymbol{x}_{i}^{1} \right) -\boldsymbol{h}_i \rVert ^2}+\sum_{j=1}^{n_m+n_2}{\lVert f^2\left( \theta _2;\boldsymbol{x}_{j}^{2} \right) -\boldsymbol{h}_j \rVert ^2},
\end{align}
\end{small}
where $\theta _1$ and $\theta _2$ denote parameters of imgNet and txtNet. Since the Euclidean distance reveals the difference between the outputs of networks and hash codes, it guides imgNet and txtNet to adjust feature extraction strategy and approximate binary outputs. Meanwhile, since these two networks has the same target space, modality gap can be further crossed on the original feature level. Therefore, the outputs of 4096-dimensional layers in imgNet and txtNet~(i.e., the 7th layer of imgNet and the 2nd layer of txtNet) are utilized to update $\boldsymbol{z}_{i}^{v}$, which are the inputs of CLE in next iteration. 
\subsubsection{Deep Manifold Hashing}
Combining Eq.~(6), Eq.~(11) and Eq.~(12), the overall objective function of our DMH can be formulated as:
\begin{small}
\begin{align}
\underset{\boldsymbol{Q}^v,\boldsymbol{Y},\boldsymbol{W},\bar{\boldsymbol{Z}}, \boldsymbol{H}, \theta _1, \theta _2}{\min}\ell _1+\ell _2+\ell _3.
\end{align}
\end{small}
\subsection{Optimization}
The overall optimization procedure of Eq.~(13) can be decomposed three parts (i.e., CLE, GBE and DAM models) with respect to modality-invariant features $\boldsymbol{Y}$, hash codes $\boldsymbol{H}$ and parameters of hash functions $\theta_1, \theta _2$ using the alternating optimization method.

\subsubsection{The solution to CLE}
The modality-invariant features $\boldsymbol{Y}$ can be obtained by alternatively optimizing variables in CLE model~(i.e., Eq.~(6)) including $\boldsymbol{Q}^v$, $\boldsymbol{W}$, $\bar{\boldsymbol{Z}}$ and $\boldsymbol{Y}$.

Specifically, the optimization procedure starts from $\boldsymbol{Q}^v$. Given $\boldsymbol{W}$, $\bar{\boldsymbol{Z}}$ and $\boldsymbol{Y}$, Eq.~(6) with respect
to $\boldsymbol{Q}^v$ is defined as:
\begin{small}
\begin{align}
\underset{\boldsymbol{Q}^v}{\min}\ell _{\boldsymbol{Q}^v}&=\frac{1}{2n}\sum_{i=1}^n{\sum_{v=1}^2{\lVert \bar{\boldsymbol{z}}_{i}^{v}-\boldsymbol{Q}^v\boldsymbol{y}_i \rVert ^2}}\\\nonumber
&s.t. \left( \boldsymbol{Q}^v \right) ^T\boldsymbol{Q}^v=I.
\end{align}
\end{small}
Setting the gradient of $\ell _{\boldsymbol{Q}^v}$ with respect to $\boldsymbol{Q}^v$ to 0 without the orthogonal constraint, we have: 
\begin{small}
\begin{align}
\sum_{i=1}^n{\left( -\bar{\boldsymbol{z}}_{i}^{v}\boldsymbol{y}_{i}^{T}+\boldsymbol{Q}^v\boldsymbol{y}_i\boldsymbol{y}_{i}^{T} \right)}=0.
\end{align}
\end{small}
The above equation can be arranged as:
\begin{small}
\begin{align}
\boldsymbol{Q}^v=\left( \sum_{i=1}^n{\bar{\boldsymbol{z}}_{i}^{v}\boldsymbol{y}_{i}^{T}} \right) \left( \sum_{i=1}^n{\boldsymbol{y}_i\boldsymbol{y}_{i}^{T}} \right) ^{-1}.
\end{align}
\end{small}
Inspired by~\cite{ding2016robust}, we directly orthogonalize the result of Eq.~(16) to satisfy the orthogonal constraint. Therefore, we can iteratively update $\boldsymbol{Q}^v$ using the orthogonal result of Eq.~(16).

Then, fix $\boldsymbol{Q}^v$, $\bar{\boldsymbol{Z}}$ and $\boldsymbol{Y}$, Eq.~(6) with respect to $\boldsymbol{W}$ reduces to:
\begin{small}
\begin{align}
\underset{\boldsymbol{W}}{\min}\ell _{\boldsymbol{W}}&=\frac{\lambda}{n}{\lVert \boldsymbol{y}_i-\sum_{j=1}^k{\boldsymbol{W}_{ij}\boldsymbol{y}_{ij}} \rVert^2}\\\nonumber &s.t.\sum_{j=1}^k{\boldsymbol{W}_{ij}}=1.
\end{align}
\end{small}
Due to the sum constraint of $\sum_{j=1}^k{\boldsymbol{W}_{ij}}$, Eq.~(17) without constraint can be written as:
\begin{small}
\begin{align}
\underset{\boldsymbol{W}}{\min}\ell _{\boldsymbol{W}}&=\frac{\lambda}{n}\,\,\left\| \sum_{j=1}^k{\boldsymbol{W}_{ij}}\boldsymbol{y}_i-\sum_{j=1}^k{\boldsymbol{W}_{ij}\boldsymbol{y}_{ij}} \right\| ^2\\\nonumber&=\frac{\lambda}{n}\,\left\| \sum_{j=1}^k{\boldsymbol{W}_{ij}\left( \boldsymbol{y}_i-\boldsymbol{y}_{ij} \right)} \right\| ^2.
\end{align}
\end{small}
The above problem can be solved by Augmented Lagrangian Multiplier~(ALM)~\cite{glowinski1989augmented}. Since $\boldsymbol{W}_i=\left(\boldsymbol{W}_{i1},\boldsymbol{W}_{i2},...,\boldsymbol{W}_{ik} \right)$ is a $k\times 1$ dimensional vector, Eq.~(18) can be translated into:
\begin{small}
\begin{align}
\underset{\boldsymbol{W}}{\min}\ell _{\boldsymbol{W}}&=\frac{\lambda}{n}\,\left\| \sum_{j=1}^k{\boldsymbol{W}_{ij}\left( \boldsymbol{y}_i-\boldsymbol{y}_{ij} \right)} \right\| ^2+\frac{\lambda}{n}\lambda^\prime\left( \boldsymbol{W}_{i}^{T}\boldsymbol{1}_k-1 \right) \\\nonumber&=\frac{\lambda}{n}\,\,\left( \sum_{j=1}^k{\boldsymbol{W}_{ij}\left( \boldsymbol{y}_i-\boldsymbol{y}_{ij} \right)} \right) ^T\left( \sum_{j=1}^k{\boldsymbol{W}_{ij}\left( \boldsymbol{y}_i-\boldsymbol{y}_{ij} \right)} \right) +\frac{\lambda}{n}\lambda^\prime\left( \boldsymbol{W}_{i}^{T}\boldsymbol{1}_k-1 \right),
\end{align}
\end{small}
where $\lambda '>0$ is a penalty parameter. To facilitate understanding, we define $\boldsymbol{B}_i=\left( \left( \boldsymbol{y}_i-\boldsymbol{y}_{i1} \right) ,\left( \boldsymbol{y}_i-\boldsymbol{y}_{i2} \right) ,...,\left( \boldsymbol{y}_i-\boldsymbol{y}_{ik} \right) \right)$ and $\boldsymbol{A}_i=\boldsymbol{B}_{i}^{T}\boldsymbol{B}_i$. Therefore, Problem~(19) can be reformulated as:
\begin{small}
\begin{align}
\underset{\boldsymbol{W}}{\min}\ell _{\boldsymbol{W}}&=\frac{\lambda}{n}\,\,\boldsymbol{W}_{i}^{T}\boldsymbol{B}_{i}^{T}\boldsymbol{B}_i\boldsymbol{W}_i+\frac{\lambda}{n}\lambda^\prime\left( \boldsymbol{W}_{i}^{T}\boldsymbol{1}_k-1 \right) \\\nonumber&=\frac{\lambda}{n}\,\,\boldsymbol{W}_{i}^{T}\boldsymbol{A}_i\boldsymbol{W}_i+\frac{\lambda}{n}\lambda^\prime\left( \boldsymbol{W}_{i}^{T}\boldsymbol{1}_k-1 \right).
\end{align}
\end{small}
Setting the gradient of $\ell_{\boldsymbol{W}}$ with respect to $\boldsymbol{W}$ to 0, we have
\begin{small}
\begin{align}
\frac{\lambda}{n}\,\,2\boldsymbol{A}_i\boldsymbol{W}_i+\frac{\lambda}{n}\lambda '\boldsymbol{1}_k=0.
\end{align}
\end{small}
Since $\sum_{j=1}^k{\boldsymbol{W}_{ij}}=1$, $\boldsymbol{W}_i$ can be represented as:
\begin{small}
\begin{align}
\boldsymbol{W}_i=\boldsymbol{A}_{i}^{-1}\boldsymbol{1}_k\left( 1_{k}^{T}\boldsymbol{A}_{i}^{-1}\boldsymbol{1}_k \right) ^{-1}.
\end{align}
\end{small}
Furthermore, based on the updated $\boldsymbol{W}$, incomplete-modality features $\boldsymbol{z}_{j}^{v}$ can be complemented as fully-paired descriptions according to Eq.~(5) and thus $\bar{\boldsymbol{Z}}$ is updated.

Finally, the optimization objective is $\boldsymbol{Y}$. Fix $\boldsymbol{Q}^v$, $\bar{\boldsymbol{Z}}$ and $\boldsymbol{W}$, Eq.~(6) with respect to $\boldsymbol{Y}$ is defined as:
\begin{small}
\begin{align}
\underset{\boldsymbol{Y}}{\min}\ell _{\boldsymbol{Y}}&=\frac{1}{2n}\sum_{v=1}^2{\lVert \bar{\boldsymbol{z}}_{i}^{v}-\boldsymbol{Q}^v\boldsymbol{y}_i \rVert ^2}+\frac{\lambda}{n}\lVert \boldsymbol{y}_i-\sum_{j=1}^k{\boldsymbol{W}_{ij}\boldsymbol{y}_{ij}} \rVert ^2+\eta\lVert \boldsymbol{y}_i \rVert ^2.
\end{align}
\end{small}
Setting the gradient of $\ell_{\boldsymbol{Y}}$ with respect to $\boldsymbol{Y}$ to 0, we have
\begin{small}
\begin{align}
\frac{1}{2n}\sum_{v=1}^2{\left( -2\left( \boldsymbol{Q}^v \right) ^T\bar{\boldsymbol{z}}_{i}^{v}+2\left( \boldsymbol{Q}^v \right) ^T\boldsymbol{Q}^v\boldsymbol{y}_i \right)}+\frac{\lambda}{n}\left( 2\boldsymbol{\boldsymbol{y}}_i-2\sum_{j=1}^k{\boldsymbol{\boldsymbol{W}}_{ij}\boldsymbol{\boldsymbol{y}}_{ij}} \right) +2\eta \boldsymbol{\boldsymbol{y}}_i=0.
\end{align}
\end{small}
The above equation can be arranged as:
\begin{small}
\begin{align}
\boldsymbol{y}_i\!=\!\left(\!\sum_{v=1}^2{\left(\left( \boldsymbol{Q}^v \right) ^T\bar{\boldsymbol{z}}_{i}^{v} \right)}+2\lambda\left( \sum_{j=1}^k{\boldsymbol{W}_{ij}\boldsymbol{y}_{ij}} \right)\!\right)\!\left( \sum_{v=1}^2{\left(\left( \boldsymbol{Q}^v \right)^T \boldsymbol{Q}^v\right)}+2\left(\lambda+\eta n\right)\boldsymbol{I} \right)^{-1}.
\end{align}
\end{small}
The detailed iteration process of CLE model is summarized in Algorithm~\ref{alg1}.
  

\begin{small}
\begin{algorithm}[!ht]
  \caption{Complemented Local Embedding~(CLE)}
  \SetKwInOut{Input}{Input}\SetKwInOut{Output}{Output}
 \Input{$\boldsymbol{Z}^{1}$, $\boldsymbol{Z}^{2}$ and $\lambda$.}
  \Output{$\boldsymbol{Y}$}
  {\bfseries Initialization:} {Randomly initialize $\boldsymbol{Q}^1$, $\boldsymbol{Q}^2$, $\boldsymbol{W}$, $\boldsymbol{Y}$ and complement the $\bar{\boldsymbol{Z}}$ of semi-paired objects with zeros.}
  
\Repeat{the estimates of $\boldsymbol{Y}$ converge}
{
Update $\boldsymbol{Q}^v$ by the orthogonal result of Eq.~(16);

Update $\boldsymbol{W}$ by Eq.~(22);

Update $\bar{\boldsymbol{Z}}$ by Eq.~(5);

Update $\boldsymbol{Y}$ by Eq.~(25);
}
\Return{$\boldsymbol{Y}$}
\label{alg1}\end{algorithm}
\end{small}

\subsubsection{The solution to GBE}
Once modality-invariant representations $\boldsymbol{Y}$ are obtained, GBE model aims to acquire hash codes $\boldsymbol{H}$ based on them. The binary $\boldsymbol{H}$ is relaxed as continuous $\hat{\boldsymbol{H}}$. Since Eq.~(11) is derivable for $\hat{\boldsymbol{H}}$, its optimization can be solved by Back-propagation algorithm~(BP). In particular, for each $\hat{\boldsymbol{h}_i}$, we first compute the following gradient:
\begin{small}
\begin{align}
\frac{\partial \ell _2}{\partial \hat{\boldsymbol{h}_i}}\!=\!{\sum\nolimits_{j=1,j\ne i}^{n}{\left( \left( \boldsymbol{S}^Y_{i,j}-\hat{\boldsymbol{S}^H_{i,j}} \right) \left( 1+D\left( \hat{\boldsymbol{h}_i},\hat{\boldsymbol{h}_j} \right) \right) ^{-1}\left( \hat{\boldsymbol{h}_i}-\hat{\boldsymbol{h}_j} \right) +2\gamma\left( \hat{\boldsymbol{h}_i}-\boldsymbol{h}_i \right) \right)}}\!,
\end{align}
\end{small}
where $\boldsymbol{h}_i$ is the result of $sign(\hat{\boldsymbol{h}_i})$. Then, we update $\hat{\boldsymbol{h}_i}$ using gradient descent algorithm. As for its corresponding hash code $\boldsymbol{h}_i$, the $sign (\cdot)$ function is applied to derive it. The detailed step by step derivations are given in Algorithm~\ref{alg2}.
\begin{small}
\begin{algorithm}[!ht]
  \caption{Global Binary Embedding~(GBE)}
  \SetKwInOut{Input}{Input}\SetKwInOut{Output}{Output}
 \Input  {$\boldsymbol{Y}$, $c$ and learning rate $\lambda_h$.}
  \Output  {$\boldsymbol{H}$}
  {\bfseries Initialization:} {Randomly initialize $\hat{\boldsymbol{H}}$.}
  
\Repeat{the estimates of $\hat{\boldsymbol{H}}$ converge}
{
Update $\hat{\boldsymbol{h}_i}$ by BP algorithm:

$\quad \hat{\boldsymbol{h}_i}\gets \hat{\boldsymbol{h}_i}-\lambda_{\hat{\boldsymbol{h}_i}} \cdot \frac{\partial \ell _2}{\partial \hat{\boldsymbol{h}_i}}$

Update $\boldsymbol{h}_i$ by the result of $sign(\hat{\boldsymbol{h}_i})$;
}
$\boldsymbol{H}=sign(\hat{\boldsymbol{H}})$;\\
\Return{$\boldsymbol{H}$}
\label{alg2}
\end{algorithm}
\end{small}
\subsubsection{The solution to DAM}
Finally, parameters of hash functions $\left\{ \theta_v\right\} _{v=1}^{2}$ are optimized with the guidance of hash codes $\boldsymbol{H}$. For $\left\{ \theta _v \right\} _{v=1}^{2}$, Eq.~(12) is derivable. Therefore, BP with mini-batch stochastic gradient descent~(mini-batch SGD) method is applied to update them. The detailed optimization procedure is summarized in Algorithm~\ref{alg3}.
\begin{small}
\begin{algorithm}[!ht]
  \caption{Deep Adaptive Mapping~(DAM)}
  \SetKwInOut{Input}{Input}\SetKwInOut{Output}{Output}
 \Input  {$\boldsymbol{X}^{1}$, $\boldsymbol{X}^{2}$, $\boldsymbol{H}$, learning rate $\lambda_1, \lambda_2$, and iteration number $T_1, T_2$.}
  \Output  {Parameters $\theta _{1}$ and $\theta _{2}$ of imgNet and txtNet, and features $\boldsymbol{Z}^{1}$, $\boldsymbol{Z}^{2}$.}
  {\bfseries Initialization:} {Initialize $\theta _{1}$ and $\theta _{2}$.}
  
\Repeat{the estimates of $\left\{ \theta _v \right\} _{v=1}^{2}$ converge}{

\For{iter=1 to $T_1$}{
Update $\theta _{1}$ by BP algorithm:

$\quad\theta _{1}\gets \theta _{1}-\lambda_1 \cdot \frac{\partial \ell _3}{\partial \theta _{1}}$
}
\For{iter=1 to $T_2$}{
Update $\theta _{2}$ by BP algorithm:

$\quad\theta _{2}\gets \theta _{2}-\lambda_2 \cdot \frac{\partial \ell _3}{\partial \theta _{2}}$
}
}
Update $\boldsymbol{Z}^{1}$ with the outputs of the 7th layer of imgNet using $\boldsymbol{X}^{1}$ as inputs;

Update $\boldsymbol{Z}^{2}$ with the outputs of the 2nd layer of txtNet using $\boldsymbol{X}^{2}$ as inputs;

\Return{$\theta _{1}$, $\theta _{2}$, $\boldsymbol{Z}^{1}$ and $\boldsymbol{Z}^{2}$}
\label{alg3}
\end{algorithm}
\end{small}
\subsubsection{Out-of-Sample Extension}
Once the overall optimization procedure of DMH is finished, the well-trained imgNet and txtNet with $sign(\cdot)$ are used to handle out-of-sample extensions from modality $v$:
\begin{small}
\begin{align}
\boldsymbol{h}_{i}^{v}=sign\left( f^v\left( \theta _{v};\boldsymbol{x}_{i}^{v} \right) \right).
\end{align}
\end{small}
\section{Experiments}
In this section, comprehensive experiments on three real-world datasets are conducted to qualitatively and quantitatively evaluate the performance of our DMH. We first introduce the datasets used for assessment and specify the experimental setting. Following this, we demonstrate that our DMH can achieve the state-of-the-art performance on both fully-paired and semi-paired unsupervised scenarios compared with other hashing-based methods. Finally, we evaluate the influence of each sub-model for the proposed model and their corresponding parameters.
\subsection{Experimental Setting}
Three public datasets for cross-modal retrieval, namely MIRFLICKR-25K~\cite{huiskes2008mir}, MS COCO~\cite{lin2014microsoft} and NUS WIDE~\cite{chua2009nus}, are used to evaluate the performance of our DMH. The detailed description and experimental setting of each dataset are shown below. 

MIRFLICKR-25K\footnote{http://press.liacs.nl/mirflickr/dlform.html} consists of 25015 images and 223635 tags, where each image is associated with several textual tags and annotated with a 24-dimensional semantic label. Same as previous work~\cite{jiang2017deep,li2018self}, 5000 images with tags appear less than 20 times are removed to avoid noises and each text is represented by a 1386-dimensional bag-of-words vector. In test phase, we randomly sample 2000 image-text pairs as query set and regard the rest as retrieval set. In training phase, 10000 pairs from the retrieval set are used for training fully-paired methods. 

MS COCO\footnote{http://cocodataset.org/\#download} is originally collected for image understanding tasks, which contains 123287 images. Each image has its corresponding text descriptions and a predefined 81-dimensional semantic label. In experiments, all images~(87081 images) with category information are included and a 2000-dimensional bag-of-words vector is used to represent text. Specifically, 5000 pairs are randomly sampled as query set and the rest 82081 pairs work as retrieval set. For training set, we randomly sample 10000 pairs from the retrieval set.

NUS WIDE\footnote{http://lms.comp.nus.edu.sg/research/NUS-WIDE.htm} is a public dataset also crawled from the Flickr website, which comprises 269648 image-tag pairs together with a 81-dimensional semantic label. Tags of one image are represented as a binary vector according to the top 1000 most-frequent tags. Consequently, some of them are all zeros, which causes an adverse impact on evaluation. To alleviate them, we remove those image-tag pairs and follow previous work~\cite{jin2018deep,lu2019efficient} to form a new dataset consisting of 182021 image-text pairs that belong to the 21 most frequent labels. In this dataset, 2100 pairs are randomly selected as query set and the rest 179921 pairs act as retrieval set. To train models, 10500 image-text pairs from retrieval set act as training set.

The proposed DMH is compared with six state-of-the-arts including CVH~\cite{kumar2011learning}, IMH~\cite{song2013inter}, LSSH~\cite{zhou2014latent}, STMH~\cite{wang2015semantic}, CMFH~\cite{ding2016large}, and FSH~\cite{liu2017cross}, where IMH can be optimized using semi-paired un-annotated data, the others are fully-paired unsupervised algorithms. To create semi-paired scenario, we reserve a certain ratio of fully-paired data in train sets, and randomly shuffle the remaining data to destroy their correspondence. In our experiments, 4096-dimensional image features from a pre-trained CNN-F network~\cite{chatfield2014return} are used and hyper-parameters of all methods are set to achieve the best performance according to their literature for fair comparison. Specifically, except the initial input features $\boldsymbol{z}_{i}^{v}$ of CLE in our DMH are 4096-dimensional pre-trained image features and original text features, the outputs of 4096-dimensional layers in imgNet and txtNet are utilized to update the similarity relations in following iterations. Meanwhile, the relevant parameters of our DMH are set as $\lambda =\text{0.1}$, $\eta =\text{0.01}$ and $\gamma =0.01$, and their sensitivity will be analysed for completeness. In the optimization phase of the sub-model GBE in our DMH, the batch size is set as 128 and solvers with different learning rates are applied (i.e., $10^{-4.5}$ for imgNet and $10^{-3.5}$ for txtNet). To make results more convincing, all experiments are repeated five times to prevent random interference and the average of results are reported. 

\subsection{Performance evaluation on fully-paired data}
In cross-modal retrieval, there are two retrieval directions: using images to query texts~($I\rightarrow T$) and using texts to query images~($T\rightarrow I$). To evaluate performance, we use Hamming ranking and hash lookup as retrieval criterion, and set the bit length at 16 bits, 32 bits, 64 bits and 128 bits. 

\subsubsection{Hamming ranking}
Hamming ranking is to sort data points in retrieval set based on their Hamming distances to the given query point. For comparison, we adopt mean average precision~(MAP) and TopN-precision curve to measure it. 

\begin{table*}
\caption{MAP comparison on MIRFLICKR-25K, MS COCO and NUS WIDE datasets. Unless particularly stated, the proportion of fully-paired data in train set is 100\%. Otherwise, the percentage is marked out after names. The best two results are marked with {\color{red} red} and {\color{blue} blue}.}
\centering
\scalebox{0.58}[0.58]{
\begin{tabular}{c |l |c c c c |c c c c |c c c c}
\hline
\multirow{2}*{Task} & \multirow{2}*{Method} & \multicolumn{4}{c}{MIRFLICKR-25K} & \multicolumn{4}{c}{MS COCO} & \multicolumn{4}{c}{NUS WIDE}\\
\cline{3-14}
        &  &16 bits & 32 bits & 64 bits & 128 bits &16 bits & 32 bits & 64 bits & 128 bits &16 bits & 32 bits & 64 bits & 128 bits\\
\hline
\multirow{9}*{$I\rightarrow T$} & CVH~\cite{kumar2011learning}	        & 0.5492 	& 0.5521    & 0.5663    & 0.5754 & 0.4832                    &	0.4440                  &	0.4685 &	0.5236 & 0.3896 &	0.3806 &	0.3746 	&0.3700\\
                                 & IMH~\cite{song2013inter}	            & 0.6155    & 0.6042 	& 0.5942    & 0.5852 & 0.5591                    &	0.5489                  &	0.5310 &	0.5428 & 0.4273 &	0.4269 &	0.4732 	&0.5047\\
                                 & LSSH~\cite{zhou2014latent}           & 0.6052 	& 0.6066    & 0.6063 	& 0.6042 & 0.5690                    &	0.5722                  &	0.5718 &	0.5721 & 0.4202 &	0.4406 &	0.4571 	&0.4591\\
                                 & STMH~\cite{wang2015semantic}         & 0.6238 	& 0.6194    & 0.6209    & 0.6247 & 0.5774                    &	0.6153                  &	0.6279 &	0.6373 & 0.4657 &	0.4963 &	0.5272 	&0.5491\\
                                 & CMFH~\cite{ding2016large}            & 0.6393 	& 0.6456 	& 0.6488 	& 0.6452 & 0.6305                    &	0.6433                  &	0.6486 & 	0.6503 & 0.4840 &	0.4919 &	0.4963 	&0.4936\\
                                 & FSH~\cite{liu2017cross}	            & 0.6099    & 0.6207    & 0.6322    & 0.6422 & 0.6047                    &	0.6270                  &	0.6424 &	0.6535 & 0.4358 &	0.4573 &	0.4689 	&0.4850\\
                                 \cline{2-14}
                                 & DMH~(50\%)                          & {\color{blue} 0.6467} 	& {\color{blue} 0.6508}     & {\color{blue} 0.6571} 	& {\color{blue} 0.6612} & {\color{blue} 0.6430} 	&   {\color{blue}  0.6575} 	&   {\color{blue} 0.6673} &   {\color{blue} 0.6736} & {\color{blue} 0.5055}& {\color{red}0.5280} & {\color{blue} 0.5343} & {\color{blue} 0.5448}\\
                                 & DMH~(100\%)                         & {\color{red} 0.6581} 	& {\color{red} 0.6671} 	& {\color{red} 0.6760} 	    & {\color{red} 0.6819} & {\color{red} 0.6520} 	    &   {\color{red}  0.6635}   &   {\color{red} 0.6736}  &   {\color{red} 	0.6767} & {\color{red} 0.5097} 	& {\color{blue} 0.5255} & {\color{red}0.5368} & {\color{red}0.5488}\\
\hline
\multirow{9}*{$T\rightarrow I$} & CVH~\cite{kumar2011learning}	        & 0.5478 	        & 0.5511 	                    & 0.5674 	& 0.5783 & 0.4815 &	0.4411 &	0.4685 &	0.5259 & 0.3949 &	0.3844 &	0.3774 	&0.3716\\
                                 & IMH~\cite{song2013inter}	            & 0.6199 	        & 0.6083 	                    & 0.5978 	& 0.5884 & 0.5617 &	0.5525 &	0.5341 &	0.5429 & 0.4340 &	0.4182 &	0.4732 	&0.5096\\
                                 & LSSH~\cite{zhou2014latent}	        & 0.5988 	        & 0.5996 	                    & 0.6016 	& 0.6011 & 0.5329 &	0.5434 &	0.5552 &	0.5586 & 0.4040 &	0.4160 &	0.4308 	&0.4354\\
                                 & STMH~\cite{wang2015semantic}	        & 0.6294 	        & 0.6279 	                    & 0.6302 	& 0.6313 & 0.5926 &	0.6106 &	0.6181 &	0.6249 & 0.4812 &	0.5002 &	0.5217 	&0.5391\\
                                 & CMFH~\cite{ding2016large}	        & 0.6381 	        & 0.6447 	                    & 0.6469 	& 0.6415 & 0.6339 &	0.6462 &	0.6518 &	0.6528 & 0.6339 &	0.6462 &	0.6518 &	0.6528\\
                                 & FSH~\cite{liu2017cross}	            & 0.6173 	        & 0.6298 	                    & 0.6425 	& 0.6539 & 0.6064 &	0.6293 &	0.6439 &	0.6545 & 0.4290 &	0.4499 &	0.4628 	&0.4771\\
                                 \cline{2-14}
                                 & DMH~(50\%)                          & {\color{blue} 0.6459} 	& {\color{blue} 0.6546} 	& {\color{blue} 0.6642} 	& {\color{blue} 0.6611} & {\color{blue} 0.6424} & {\color{blue} 0.6593} & {\color{blue} 0.6666}  & {\color{blue} 0.6756} & {\color{blue} 0.5188}& {\color{blue}0.5456}& {\color{blue} 0.5630}   & {\color{blue} 	0.5750}\\
                                 & DMH~(100\%)                         & {\color{red} 0.6604} & {\color{red} 0.6650} 	& {\color{red} 0.6752} 	    & {\color{red} 0.6787} & {\color{red} 0.6588}  & {\color{red} 0.6669}  & {\color{red} 0.6767} 	 & {\color{red} 0.6823} & {\color{red} 0.5253} 	& {\color{red} 0.5611} & {\color{red} 0.5642} & {\color{red} 0.5797}\\ 
\hline
\end{tabular}
}
\label{tab1}
\end{table*}

MAP is the most widely used criteria metrics to measure retrieval accuracy~(more details are introduced in~\cite{ding2014collective}). To reflect overall property of rankings, the size of retrieval set is used as the retrieval radius of MAP. In Table~\ref{tab1}, we report the MAP results of all baselines and our DMH on MIRFLICKR-25K, MS COCO and NUS WIDE datasets, respectively. One should note that the train data for our DMH~(100\%) and competitors are fully-paired. To further demonstrate the superiority of our DMH, we also train our DMH with 50\% fully-paired data and 50\% semi-paired data~(i.e., DMH~($50\%$)). From Table~\ref{tab1}, we have the following observations. Firstly, our DMH outperforms all the baseline methods with different code lengths on different retrieval directions, even if half of the training data is unpaired. Compared with the best fully-paired baseline CMFH, our DMH~($50\%$) achieves absolute increases of $0.92\%/1.36\%$, $1.71\%/1.48\%$ and $3.67\%/4.66\%$ on three datasets, which demonstrates its superiority. Secondly, CCA has the worst accuracy, while others earn higher MAP. The reason for this phenomenon is that CCA focuses on modeling statistical values and ignores the preservation of similarity structure, which is the starting point of other methods. Thirdly, by comparing LSSH, STMH, CMFH, FSH to IMH, we can find that fully-paired object level encoding strategy usually achieves better performance. It is caused by that unified hash codes are learned based on common characteristics of fully-paired data, and thus avoids the disruption of modality gap in train set encoding procedure. These experimental results demonstrate that unified similarity preserving hash code is more compatible for semi-paired unsupervised hashing, which partly answers for the efficiency of our DMH. Furthermore, compared with fully-paired hashing methods like CMFH and FSH, our DMH also owns the concurrent consideration of global similarity, as well as the introduction of deep neural networks.

\begin{figure}[htbp]
\centering
\subfigure[]{
\begin{minipage}[t]{0.32\linewidth}
\centering
\includegraphics[width=1.55in]{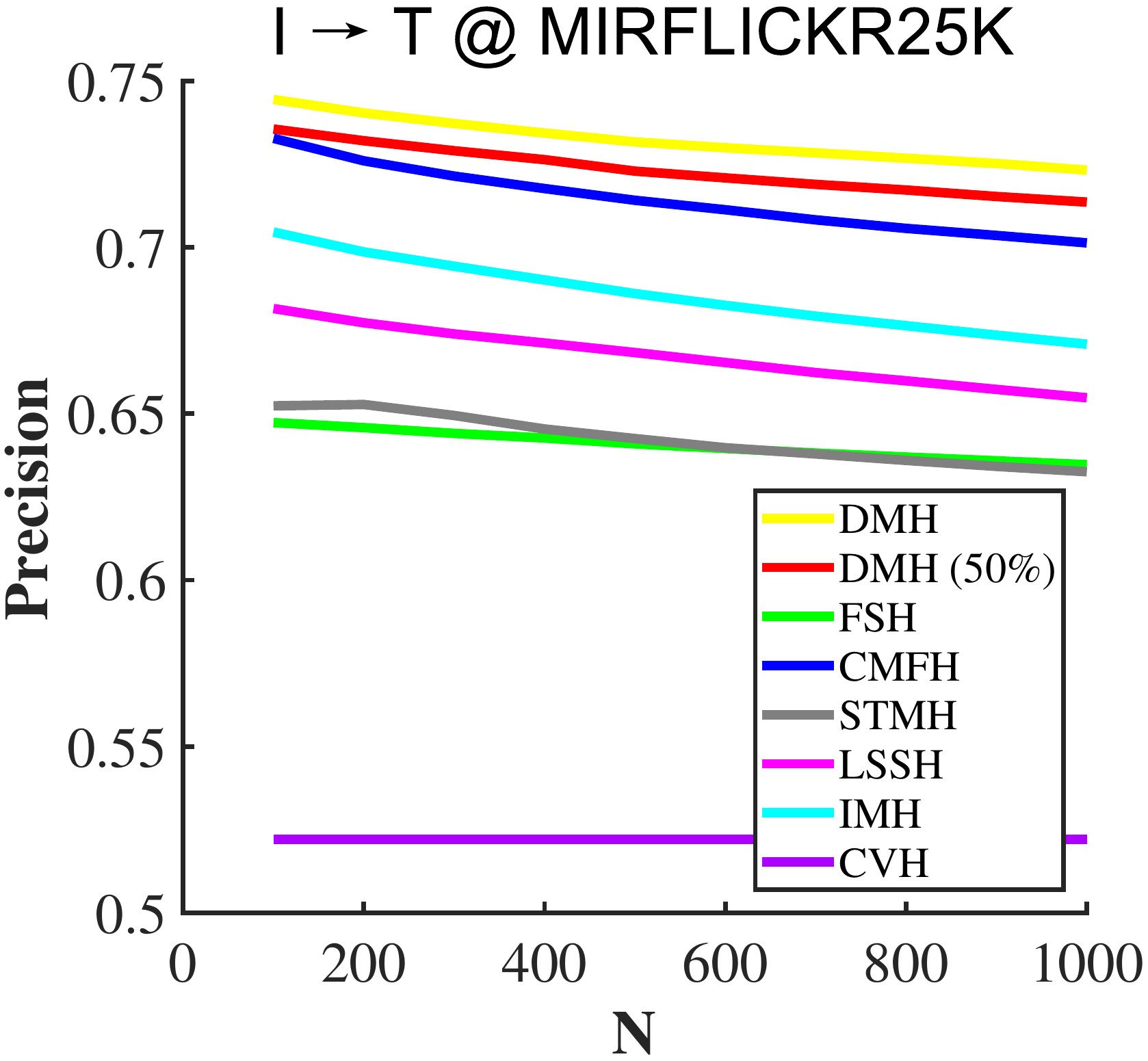}
\end{minipage}%
}%
\subfigure[]{
\begin{minipage}[t]{0.32\linewidth}
\centering
\includegraphics[width=1.55in]{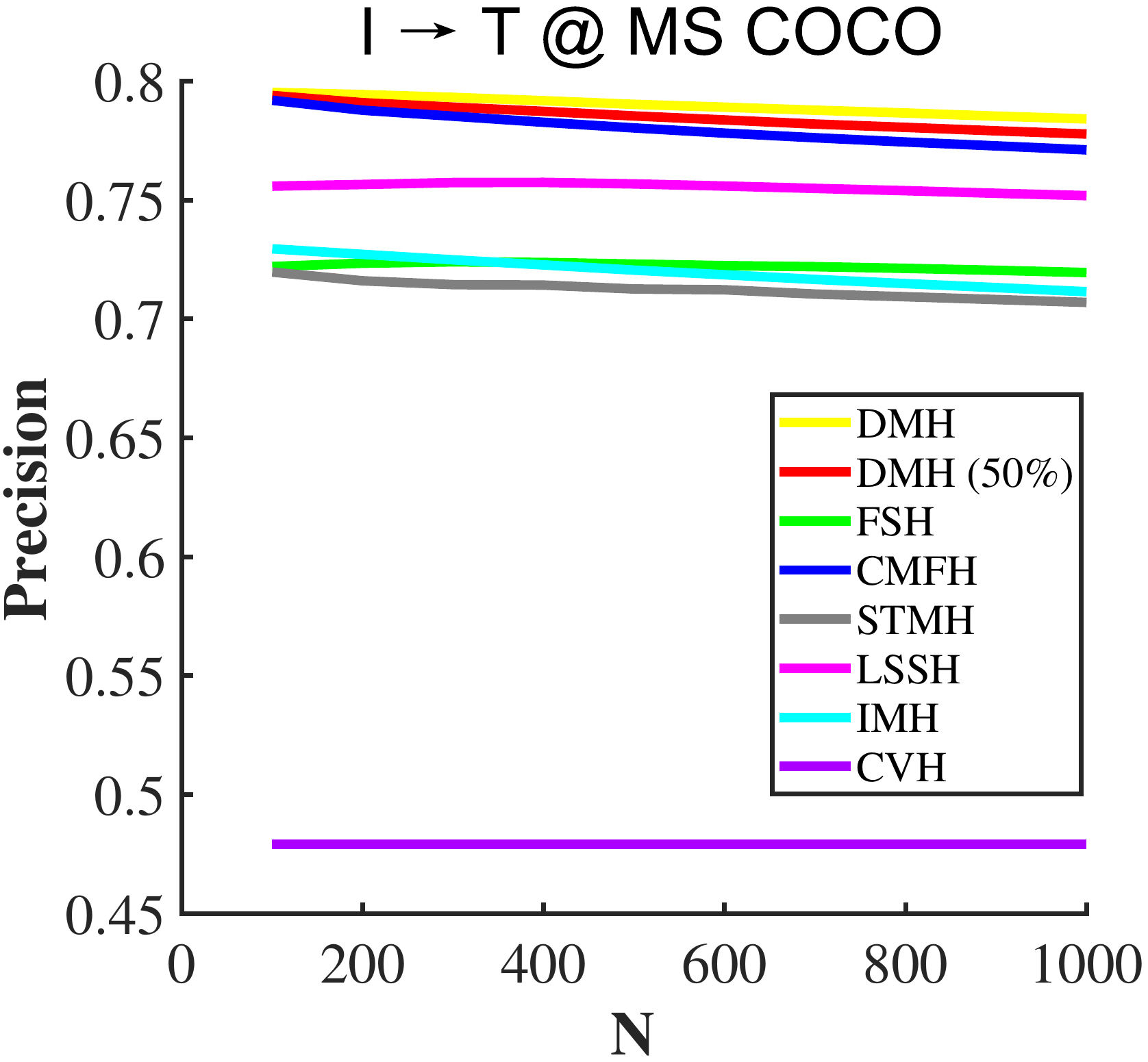}
\end{minipage}%
}%
\subfigure[]{
\begin{minipage}[t]{0.32\linewidth}
\centering
\includegraphics[width=1.55in]{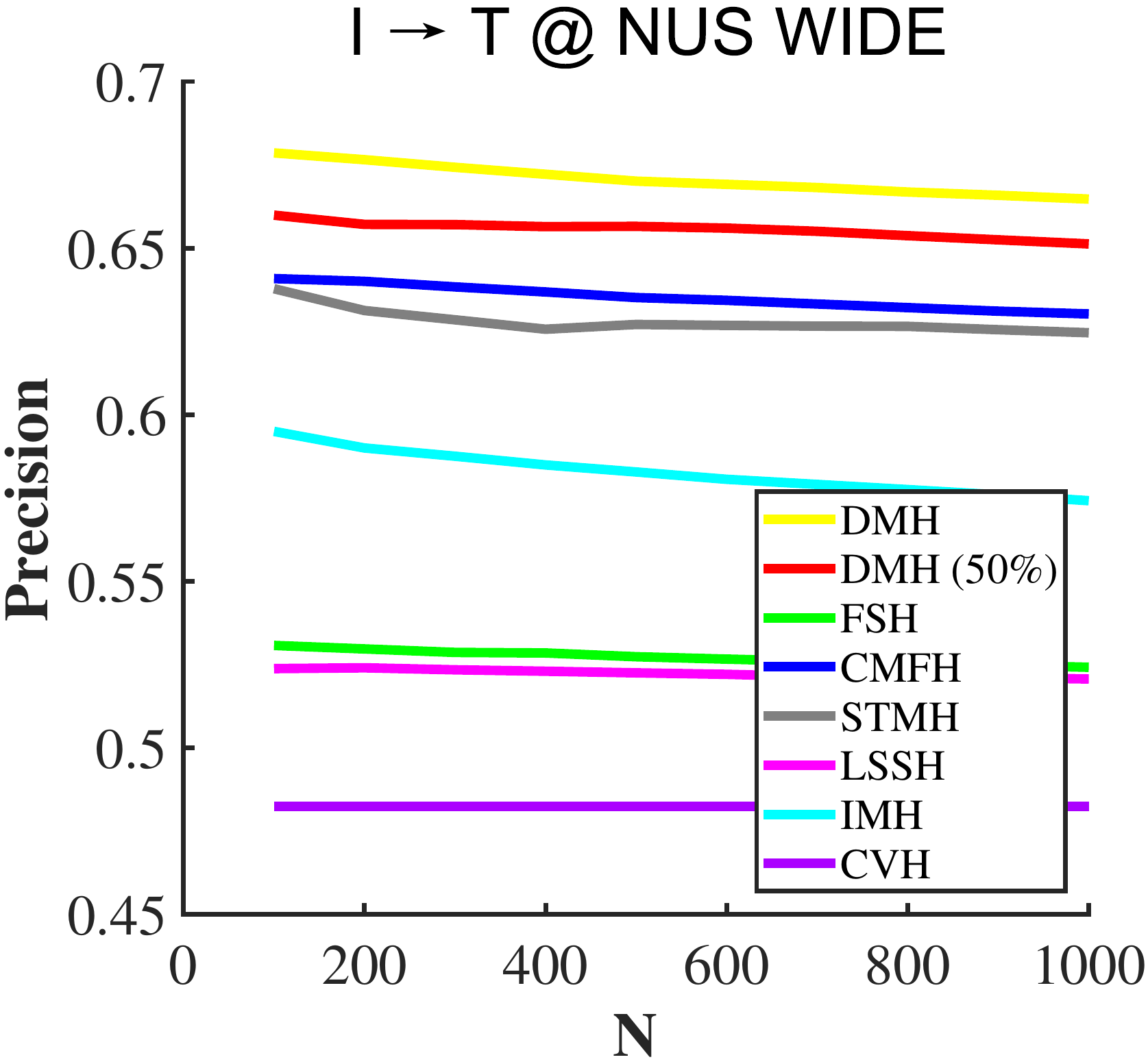}
\end{minipage}%
}%

\subfigure[]{
\begin{minipage}[t]{0.32\linewidth}
\centering
\includegraphics[width=1.55in]{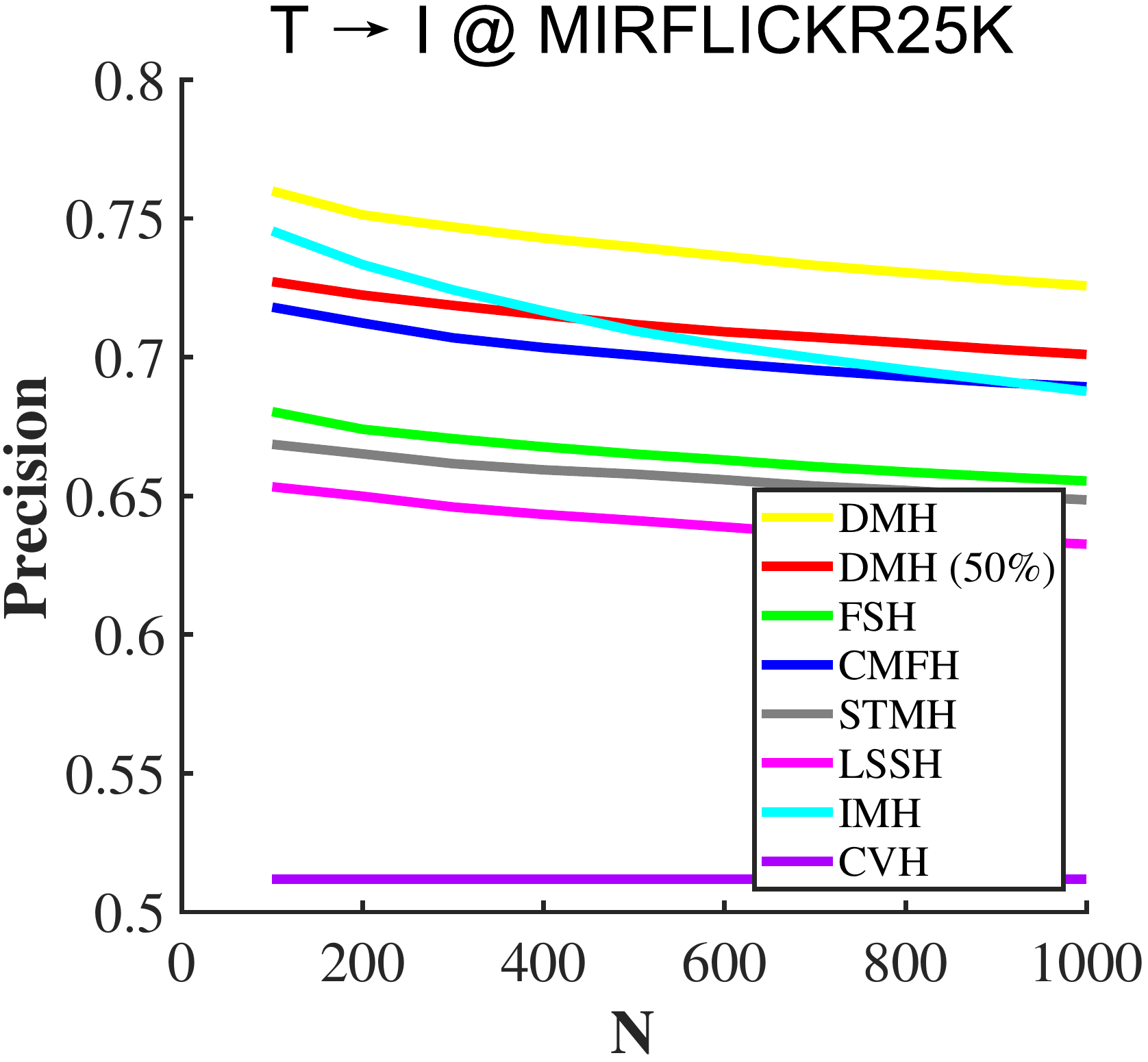}
\end{minipage}%
}%
\subfigure[]{
\begin{minipage}[t]{0.32\linewidth}
\centering
\includegraphics[width=1.55in]{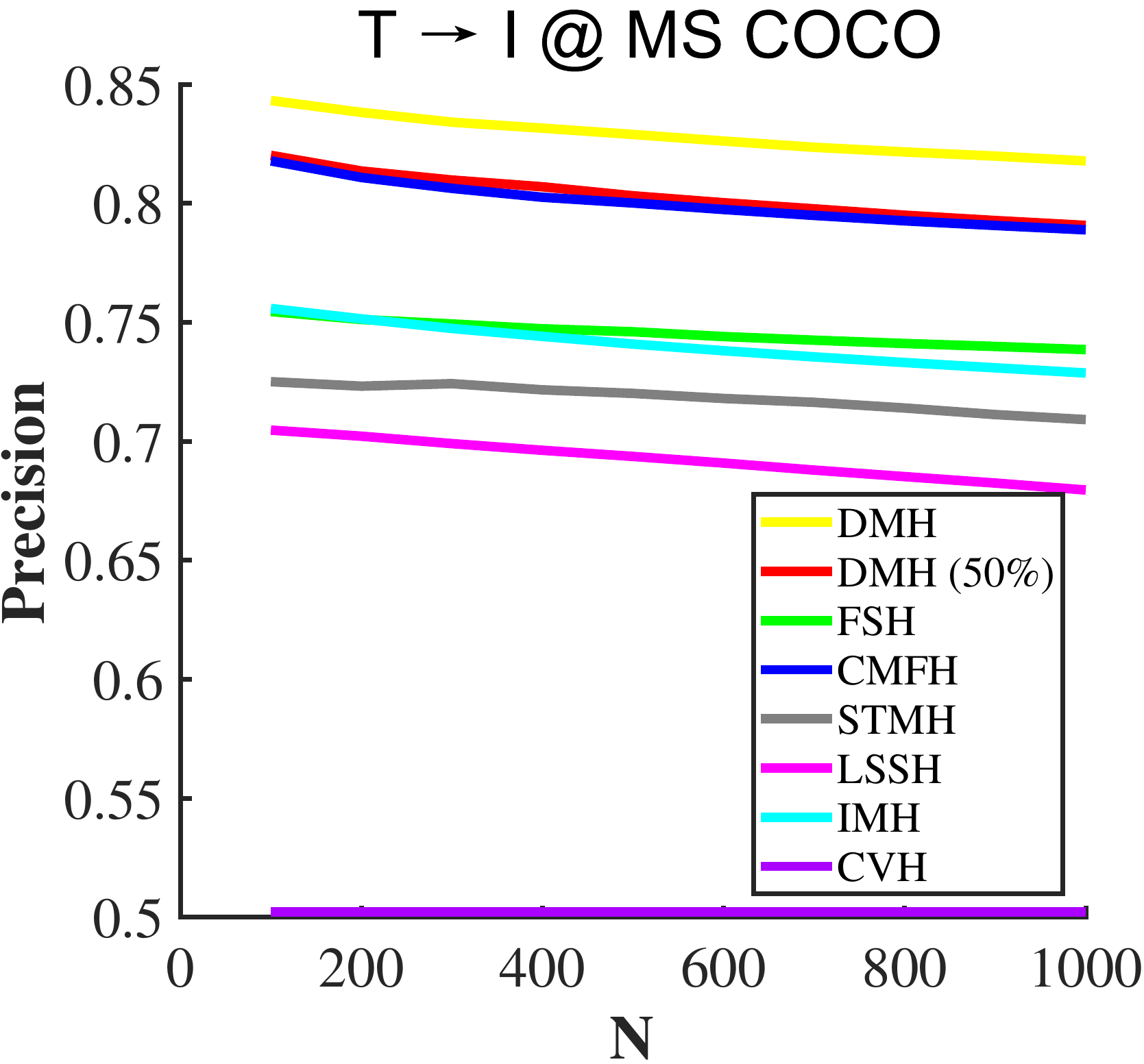}
\end{minipage}%
}%
\subfigure[]{
\begin{minipage}[t]{0.32\linewidth}
\centering
\includegraphics[width=1.55in]{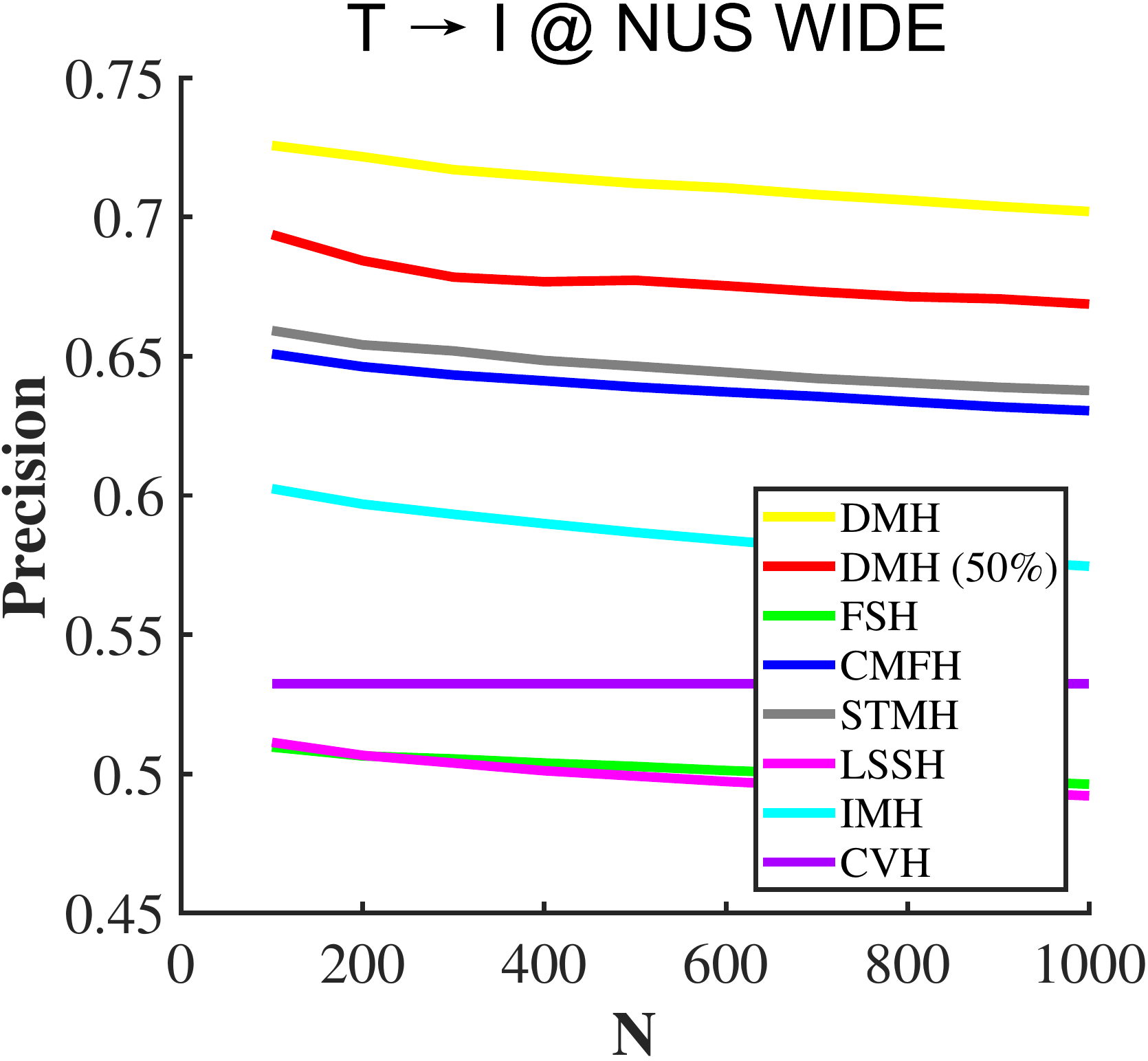}
\end{minipage}%
}%
\centering
\caption{TopN-precision curves on MIRFLICKR-25K, MS COCO and NUS WIDE datasets with code length 16. Unless particularly stated, the proportion of fully-paired data in train set is 100\%. Otherwise, the percentage is marked out after names.}
\label{fig2}
\end{figure}

In retrieval applications, we may give more attention to the precision of top-n retrieval results more than the sort quality of database. Hence, we plot TopN-precision curves to reflect the relation of precision with the number of retrieved instances on MIRFLICKR-25K, MS COCO and NUS WIDE datasets in Figure~\ref{fig2}. It can be seen that our DMH still keeps the highest precision, which reveals the advance of our methods again. Furthermore, it also reflects the efforts of our DMH in preserving local similarity, as well as the MAP comparison embodies the global characters of our DMH.

\subsubsection{Hash lookup}
Hash lookup aims to return data points in radius of a certain Hamming distance to the given query point. We use precision-recall~(PR) curve to assess its accuracy. 

\begin{figure}[htbp]
\centering
\subfigure[]{
\begin{minipage}[t]{0.32\linewidth}
\centering
\includegraphics[width=1.55in]{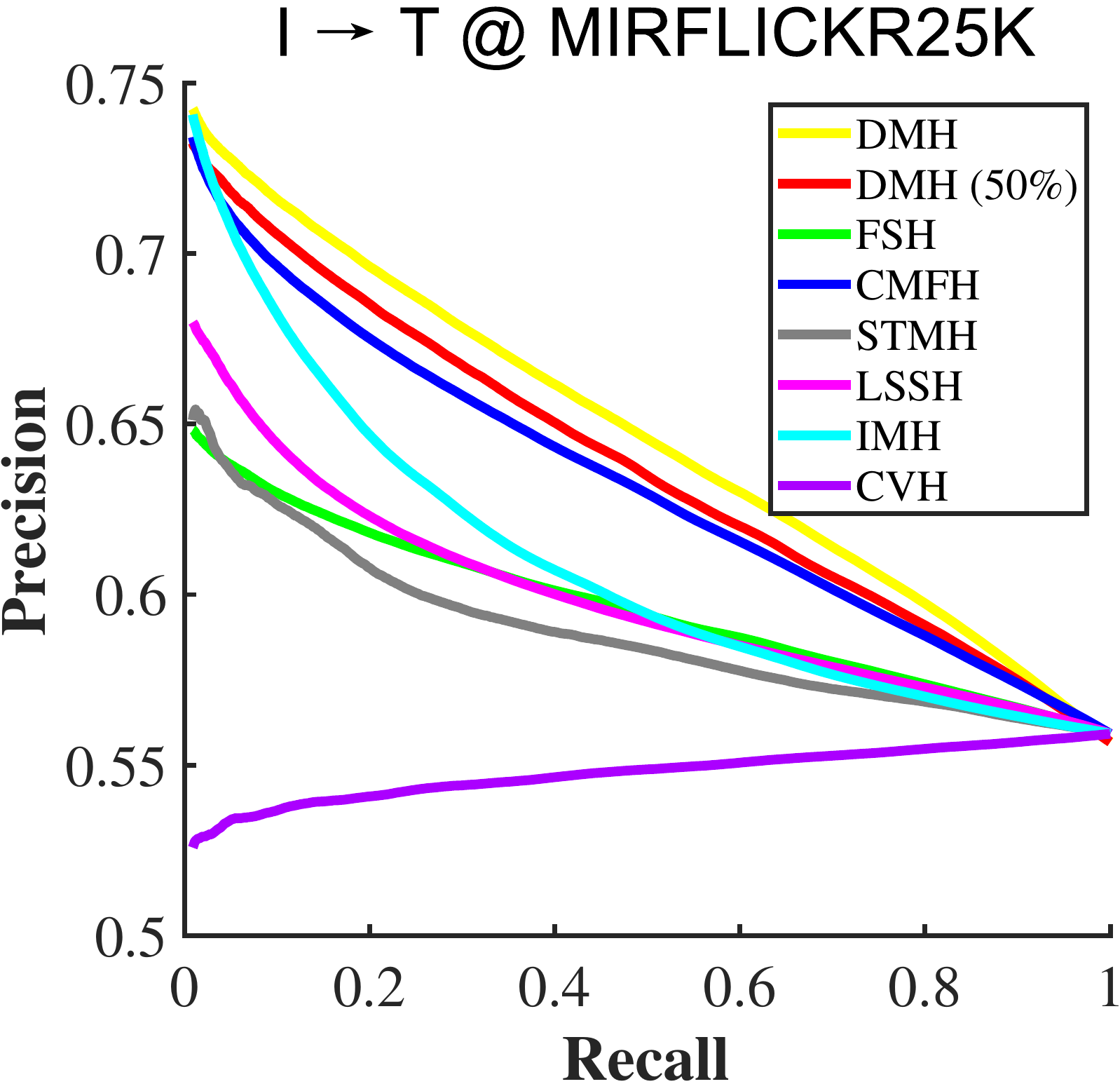}
\end{minipage}%
}%
\subfigure[]{
\begin{minipage}[t]{0.32\linewidth}
\centering
\includegraphics[width=1.55in]{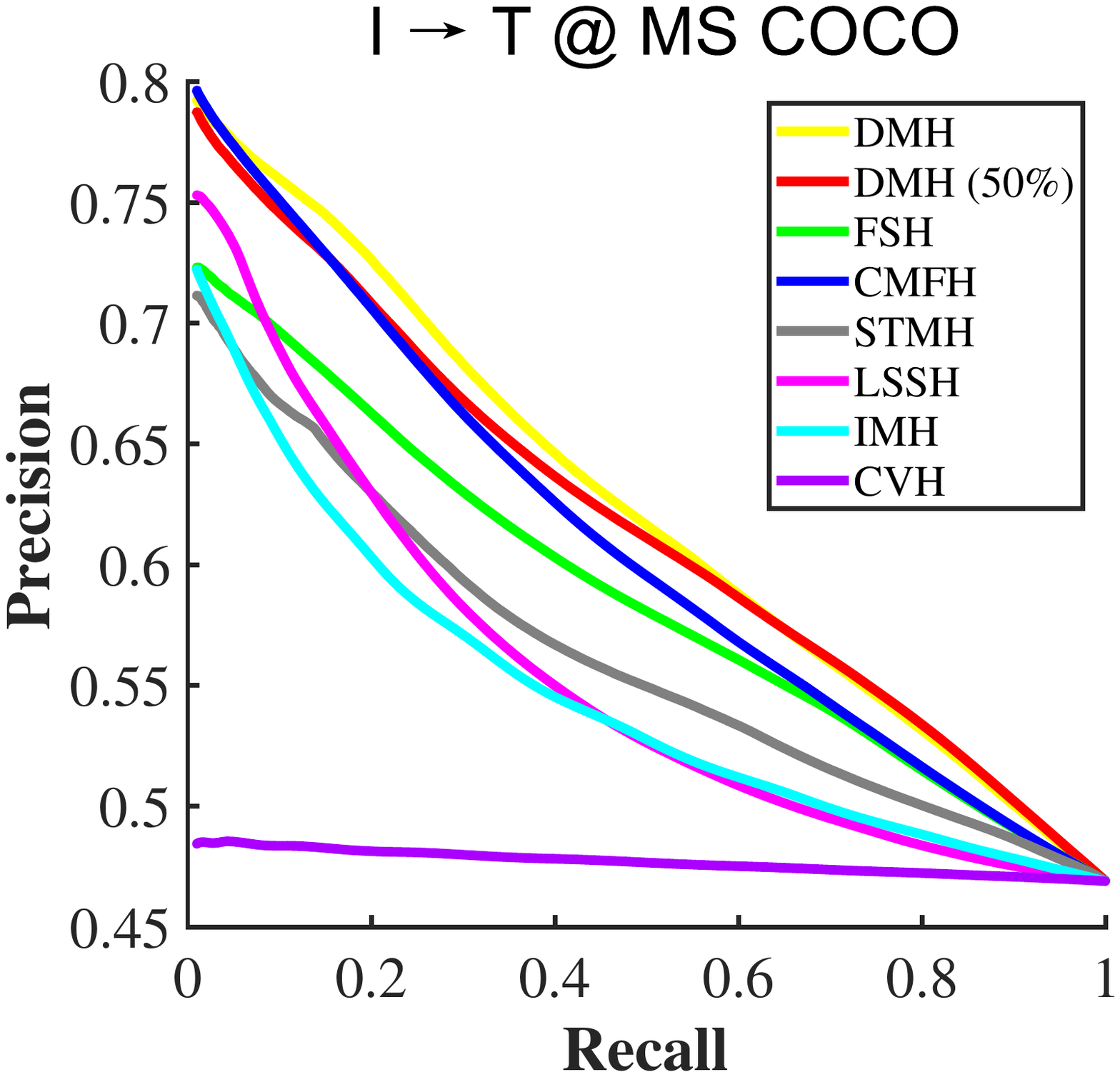}
\end{minipage}%
}%
\subfigure[]{
\begin{minipage}[t]{0.32\linewidth}
\centering
\includegraphics[width=1.55in]{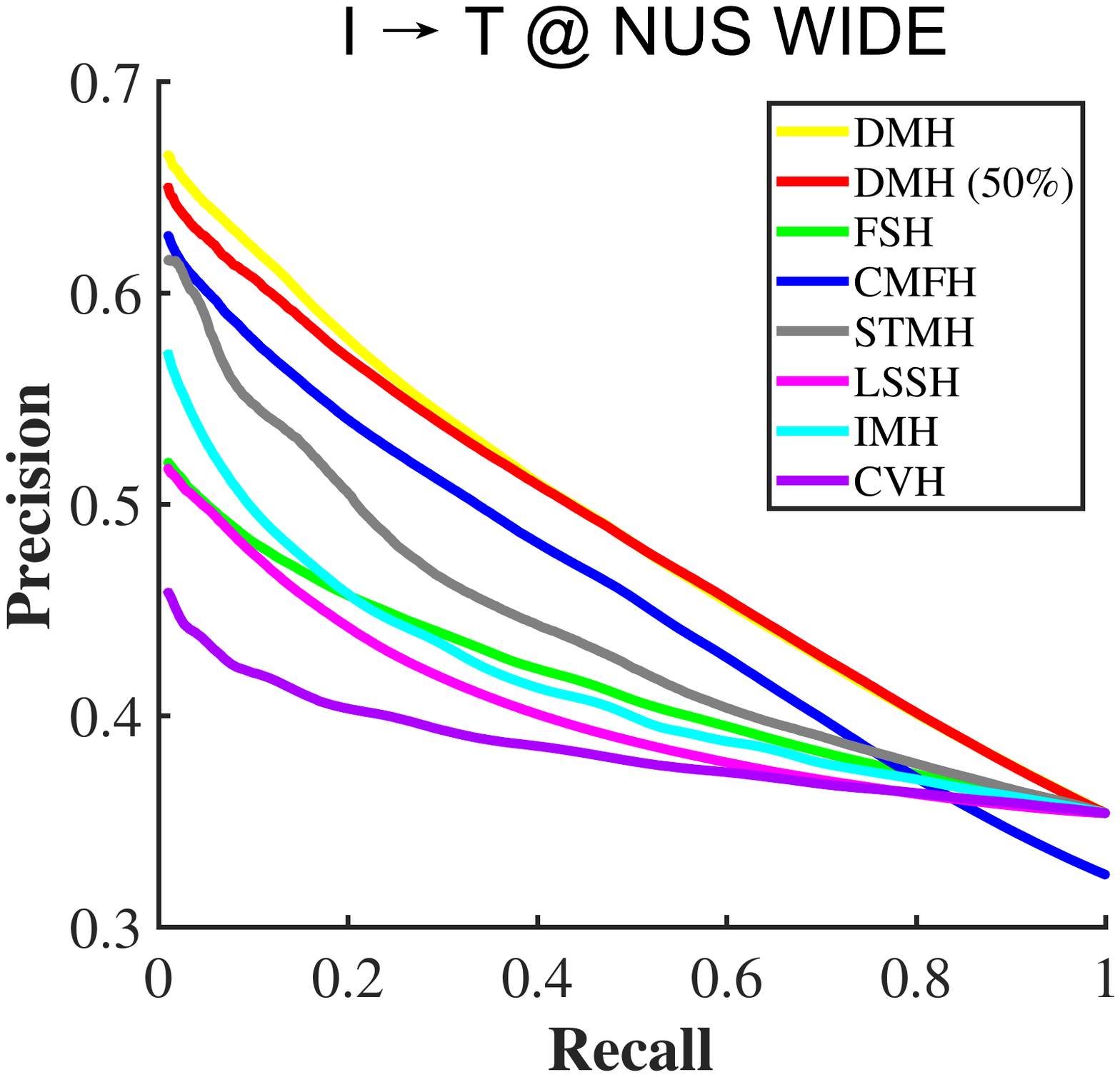}
\end{minipage}%
}%

\subfigure[]{
\begin{minipage}[t]{0.32\linewidth}
\centering
\includegraphics[width=1.55in]{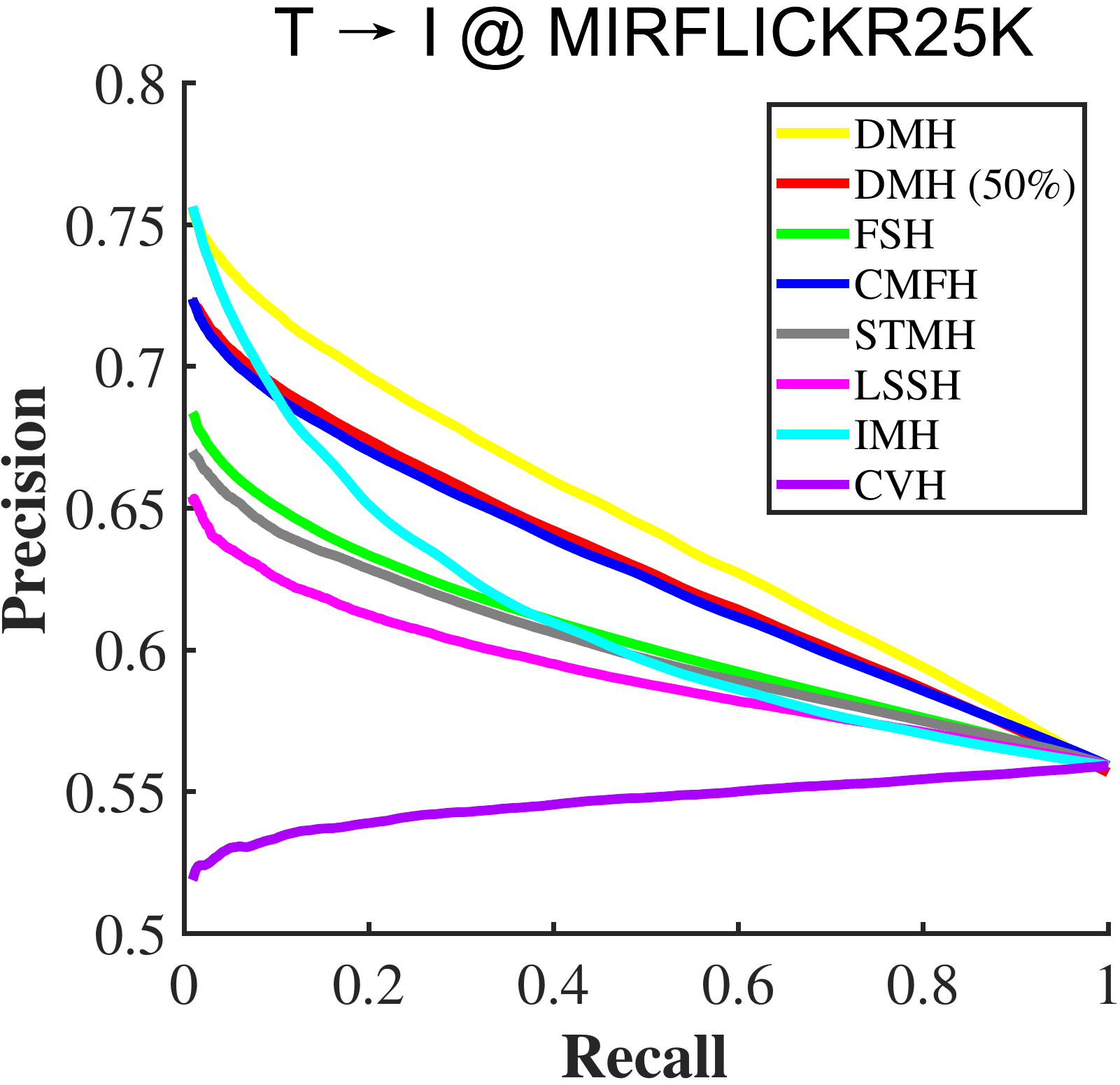}
\end{minipage}%
}%
\subfigure[]{
\begin{minipage}[t]{0.32\linewidth}
\centering
\includegraphics[width=1.55in]{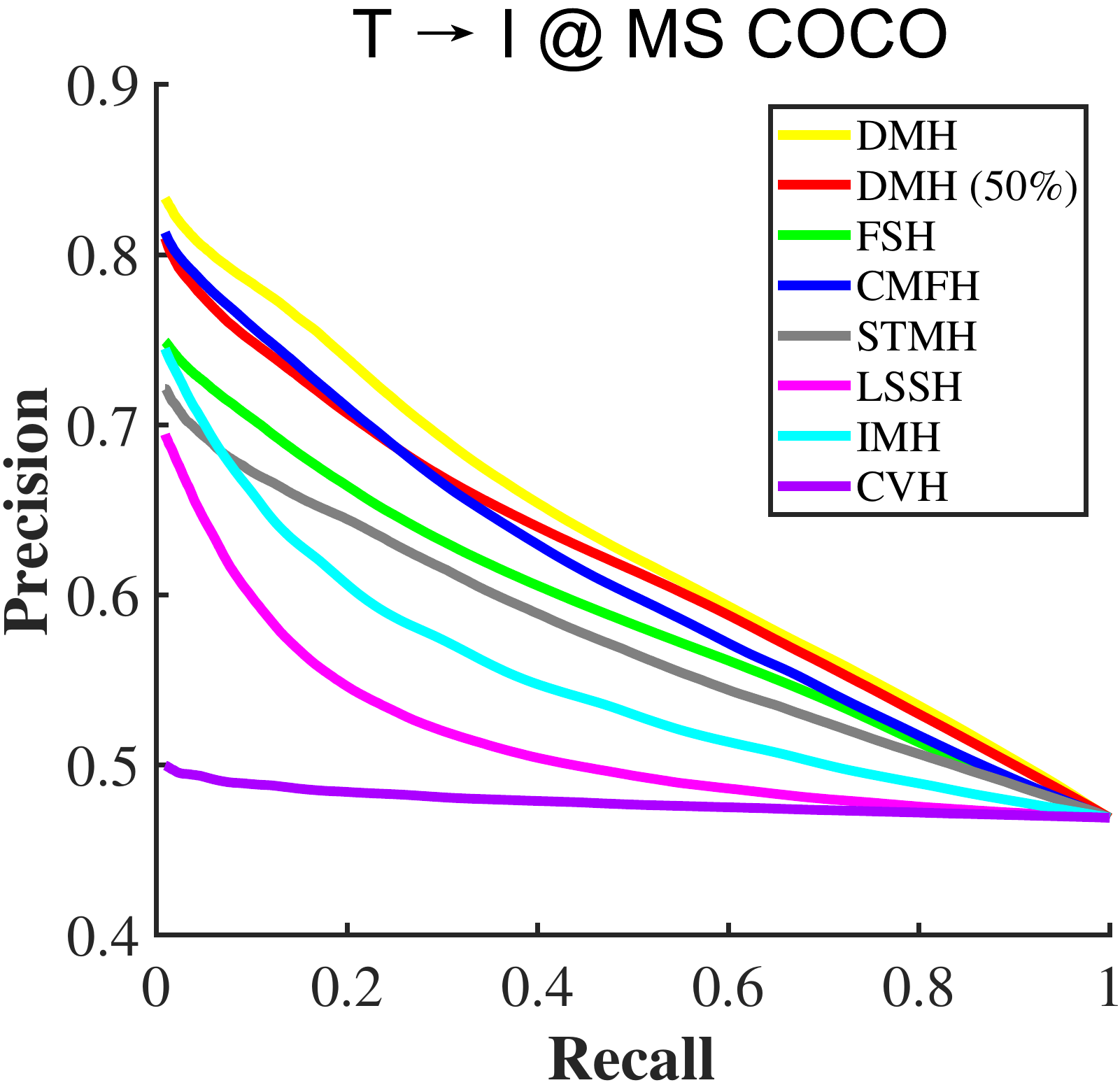}
\end{minipage}%
}%
\subfigure[]{
\begin{minipage}[t]{0.32\linewidth}
\centering
\includegraphics[width=1.55in]{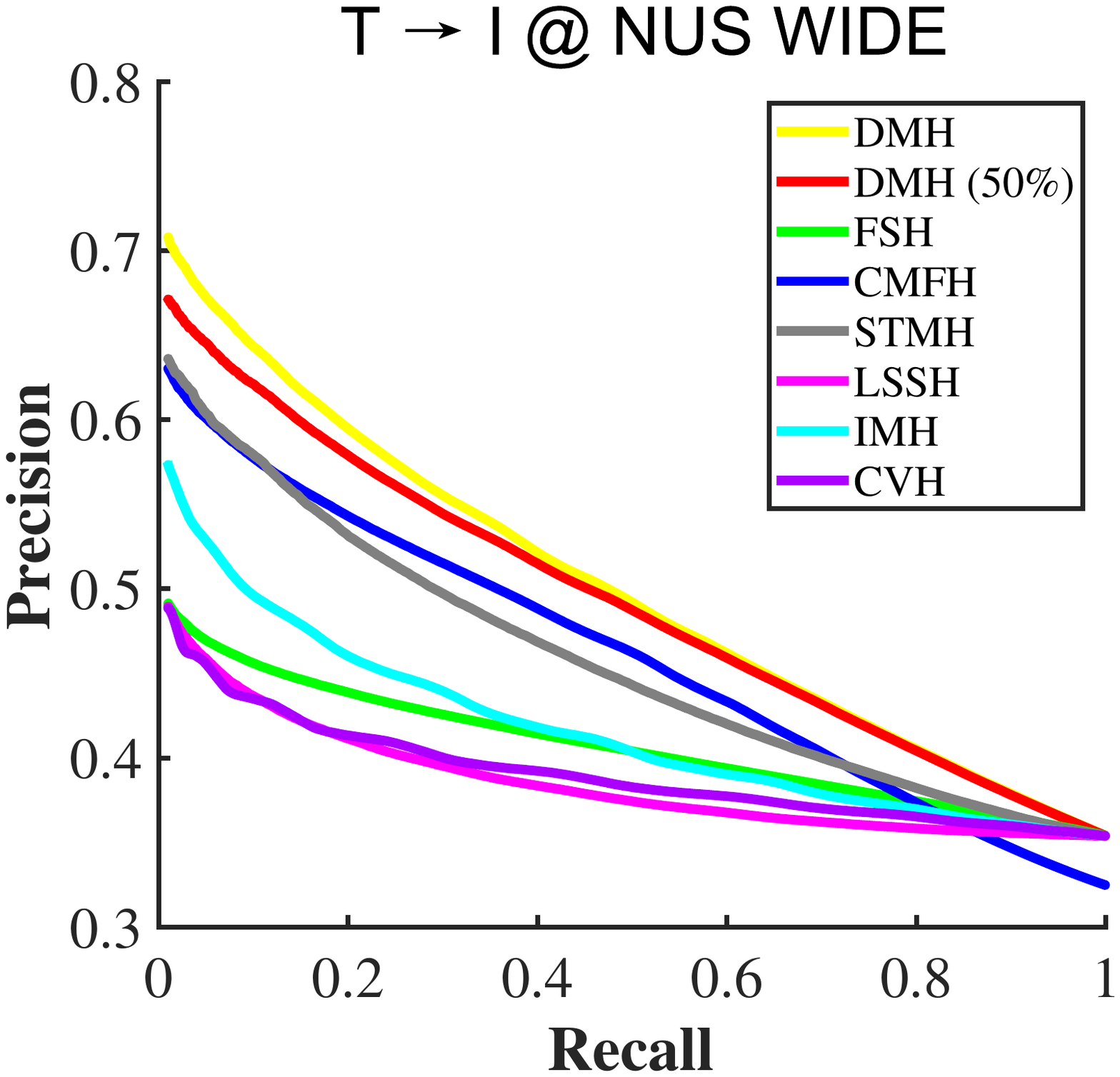}
\end{minipage}%
}%
\centering
\caption{PR curves on MIRFLICKR-25K, MS COCO and NUS WIDE datasets with code length 16. Unless particularly stated, the proportion of fully-paired data in train set is 100\%. Otherwise, the percentage is marked out after names.}
\label{fig3}
\end{figure}

The PR curves on all three datasets are plotted by varying hamming radius from 0 to 16 in Figure~\ref{fig3}, which reflects precision at different recall levels. The curves corresponding to our DMH locate higher than other curves on the whole. These phenomenons prove that our DMH achieves the state-of-the-art efficiency in Hash lookup as like its performance in Hamming ranking on fully-paired unsupervised cross-modal retrieval.

\subsection{Influence of semi-paired data}

\begin{figure}[htbp]
\centering
\subfigure[]{
\begin{minipage}[t]{0.48\linewidth}
\centering
\includegraphics[width=1.55in]{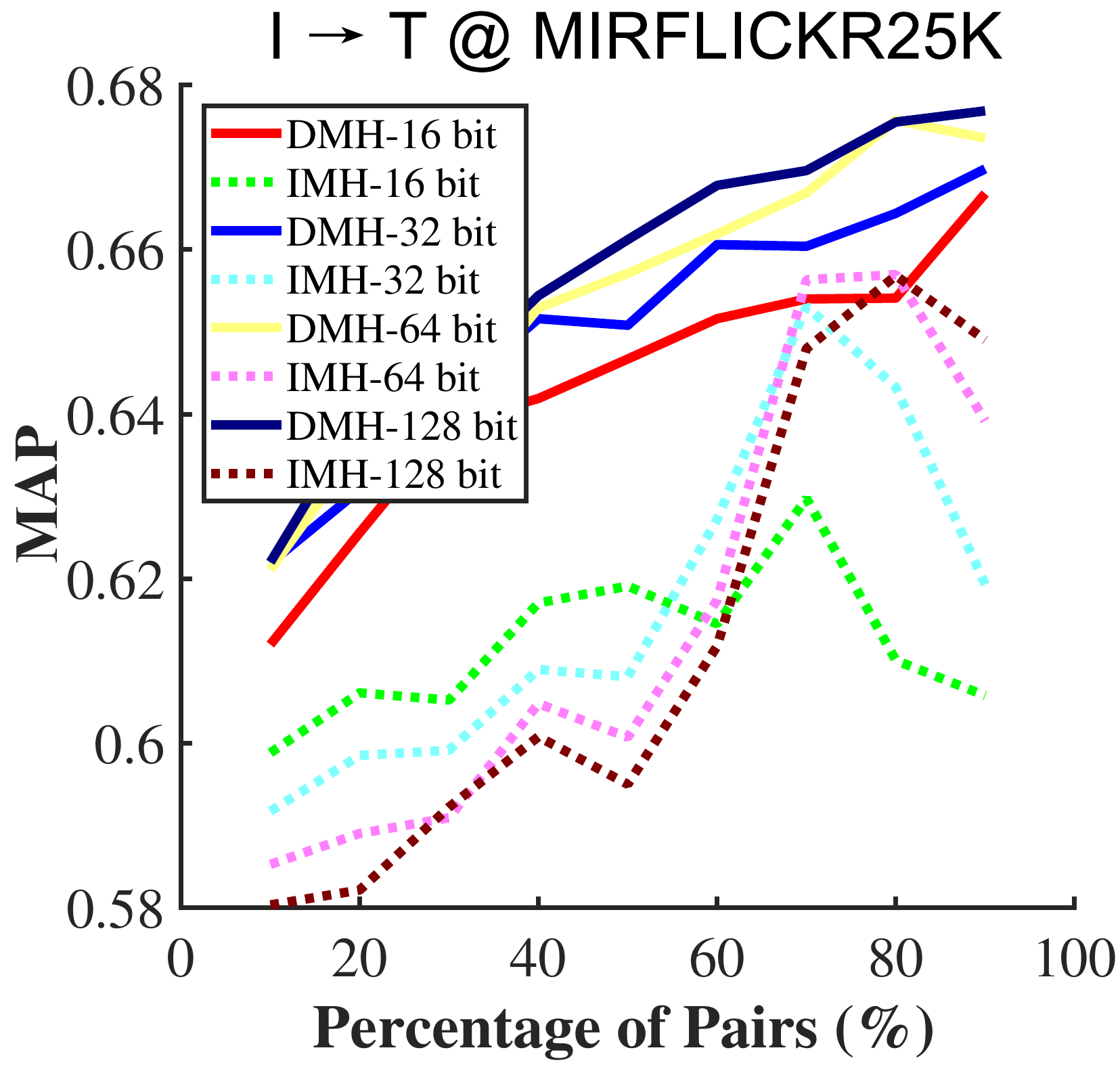}
\end{minipage}%
}%
\subfigure[]{
\begin{minipage}[t]{0.48\linewidth}
\centering
\includegraphics[width=1.55in]{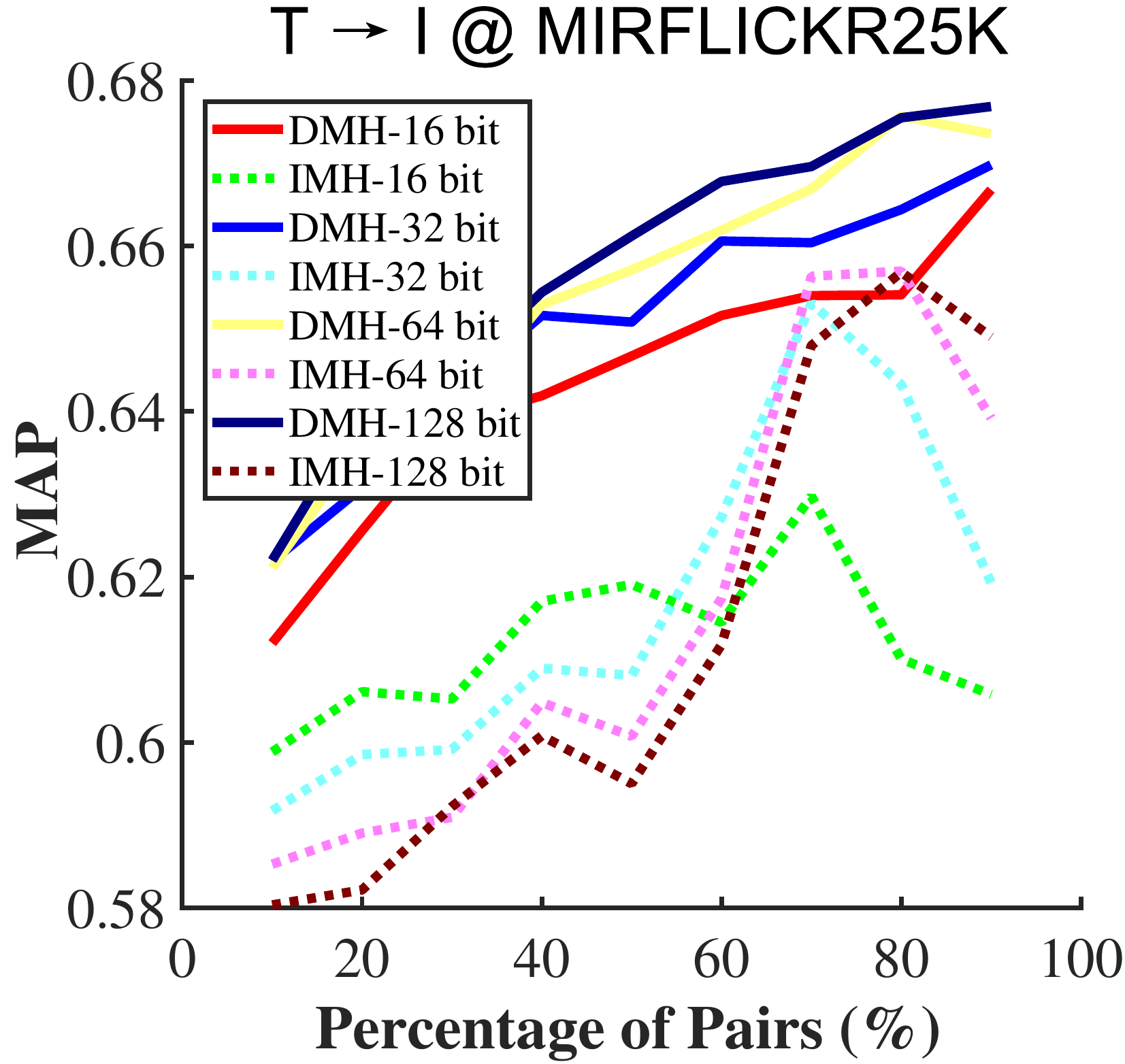}
\end{minipage}%
}%

\centering
\caption{MAP results with different pairs at different code lengths on MIRFLICKR-25K dataset.}
\label{fig4}\end{figure}

We then evaluate the performance of our DMH to demonstrate its superiority on semi-paired data. Since IMH~\cite{song2013inter} can also utilize such semi-paired data, it is selected for comparison. We vary the proportion of fully-paired data in train set from 10\% to 90\%, and report the MAP with different bit lengths for cross-modal retrieval tasks in Figure~\ref{fig4}. As can be seen, the overall variation trend of retrieval accuracy is positive with the increase of fully-paired data, which is credited with the introduction of more common characteristics. Meanwhile, our DMH outperforms IMH with significant margins across different bits. 

\begin{table*}
\caption{MAP with the retrieval radius as 50~(MAP@50) comparison on MIRFLICKR-25K dataset. The proportion of fully-paired data in train set is 50\%. The results of alternative models are reported from the reference papers and the best results are marked with {\color{red} red}.}
\centering
\scalebox{0.58}[0.58]{
\begin{tabular}{c |l |c c c c}
\hline
\multirow{2}*{Task} & \multirow{2}*{Method} & \multicolumn{4}{c}{Code Length}\\
\cline{3-6}
        &  &16 bits & 32 bits & 64 bits & 128 bits \\
\hline
\multirow{3}*{$I\rightarrow T$}  & SPDH~\cite{shen2016semi}	    & 0.6490 	& 0.6570 	& 0.6590 	& 0.6610 \\
                                 & SADCH~\cite{wang2020semi}	& 0.7160 	& 0.7690 	& 0.7810 	& 0.7930 \\
                                 \cline{2-6}
                                 & DMH     & {\color{red} 0.7864} 	& {\color{red} 0.8029} 	& {\color{red} 0.8201} 	& {\color{red} 0.8264} \\
\hline
\multirow{3}*{$T\rightarrow I$}  & SPDH~\cite{shen2016semi}	    & 0.6310 	& 0.6410 	& 0.6640 	& 0.6830 \\
                                 & SADCH~\cite{wang2020semi}	& 0.7210 	& 0.7230 	& 0.7230 	& 0.7270 \\
                                 \cline{2-6}
                                 & DMH     & {\color{red} 0.7949}	& {\color{red} 0.8021}	& {\color{red} 0.8255}	& {\color{red} 0.8330} \\
\hline
\end{tabular}
}
\label{tab7}
\end{table*}

To further verify the novelty of our DMH, we follow the experiment settings used in recently-released SPDH~\cite{shen2016semi} and SADCH~\cite{wang2020semi}, and report the experiment results in Table~\ref{tab7} where our DMH achieves a large performance gain due to the fully-paired object level encoding strategy and adaptive feature extraction ability. In addition, we comprehensively evaluate our DMH in 50\% fully-paired training data scenarios using Hamming ranking and Hash lookup. Experimental results are presented in Table~\ref{tab1}, Figure~\ref{fig2} and Figure~\ref{fig3} respectively, where values in parentheses stand for the percentage of fully-paired data~(\%). To our surprise, the proposed DMH can utilize 50\% fully-paired train set to achieve higher performances than fully-paired baselines on all cases, which we argue can be attributed to the consideration of global geometry and feature extraction. The above results further validate that the proposed DMH can handle semi-paired unsupervised cross-modal retrieval comprehensively and accurately.

\subsection{Ablation study}

\begin{table*}
\caption{MAP comparison of our DMH and its variants on MIRFLICKR-25K dataset with 50\% fully-paired data in train set. The best results are marked with {\color{red} red}.}
\centering
\scalebox{0.58}[0.58]{
\begin{tabular}{c |l |c c c c}
\hline
\multirow{2}*{Task} & \multirow{2}*{Method} & \multicolumn{4}{c}{Code Length}\\
\cline{3-6}
        &  &16 bits & 32 bits & 64 bits & 128 bits \\
\hline
\multirow{4}*{$I\rightarrow T$}  & DMH-ZERO~(50\%)	        & 0.6053 	& 0.6152 	& 0.6232 	& 0.6348 \\
                                 & DMH-PCA~(50\%)	        & 0.5968 	& 0.5878 	& 0.5825 	& 0.5757 \\
                                 & DMH-FIX~(50\%)         & 0.6323 	& 0.6410 	& 0.6522 	& 0.6565 \\
                                 \cline{2-6}
                                 & DMH~(50\%)      & {\color{red} 0.6467} 	    & {\color{red} 0.6508} 	    & {\color{red} 0.6571} 	& {\color{red} 0.6612}\\
\hline
\multirow{4}*{$T\rightarrow I$}  & DMH-ZERO~(50\%)	        & 0.5964 	& 0.6080 	& 0.6144 	& 0.6201 \\
                                 & DMH-PCA~(50\%)	        & 0.5977 	& 0.5898 	& 0.5836 	& 0.5763 \\
                                 & DMH-FIX~(50\%)         & 0.6330 	& 0.6415 	& 0.6525 	& 0.6553 \\
                                 \cline{2-6}
                                 & DMH~(50\%)      & {\color{red} 0.6459} 	& {\color{red} 0.6546} 	    & {\color{red} 0.6642} 	& {\color{red} 0.6611}\\
\hline
\end{tabular}
}
\label{tab8}
\end{table*}

To verify the effectiveness of utilizing Complemented Local Embedding~(CLE), Global Binary Embedding~(GBE) and Deep Adaptive Mapping~(DAM) in our DMH method, three variants including ``DMH-ZERO", ``DMH-PCA" and ``DMH-FIX" are designed as baselines. The proportion of fully-paired data in train set is 50\%. Table~\ref{tab8} reports their MAP results on MIRFLICKR-25K.

DMH-ZERO denotes the variant where the sub-model CLE complements incomplete samples with zeros instead of Eq.17, and other modules are still the same as those in our DMH. Therefore, DMH-ZERO can not perform incomplete sample complement during training procedure. And the comparison between DMH-ZERO and DMH in Table~\ref{tab8} validates that encoding on semi-paired level impedes the extraction of modality-invariant features and results in the residue of modality gap. DMH-PCA is built by replacing the GBE in our DMH with Principal Component Analysis~(PCA)\cite{turk1991face}. Since PCA focuses on 
statistical properties to conduct dimension reduction, DMH-PCA simulates the neglect of global data consistency. The results of DMH-PCA and DMH in Table~\ref{tab8} demonstrate that exploring global data consistency indeed boost the retrieval performance. DMH-FIX represents the variant where DMH just relies on original input features and no longer update features with the sub-model DAM. In DMH-FIX, the feature extraction procedure and hash code encoding procedure are mutually independent, results in that their features are not self-adapting. In Table~\ref{tab8}, our DMH can achieve higher performance than DMH-FIX. It indicates that the self-adapting feature extraction ability is important in hashing.

From these comparisons and analysis, we can see that our method can conduct more accurately cross-modal retrieval than other variants when using the Divide-and-Conquer strategy and proposed modules.

\subsection{Parameter analysis and convergence testing}

\begin{figure}[htbp]
\centering
\subfigure[]{
\begin{minipage}[t]{0.32\linewidth}
\centering
\includegraphics[width=1.55in]{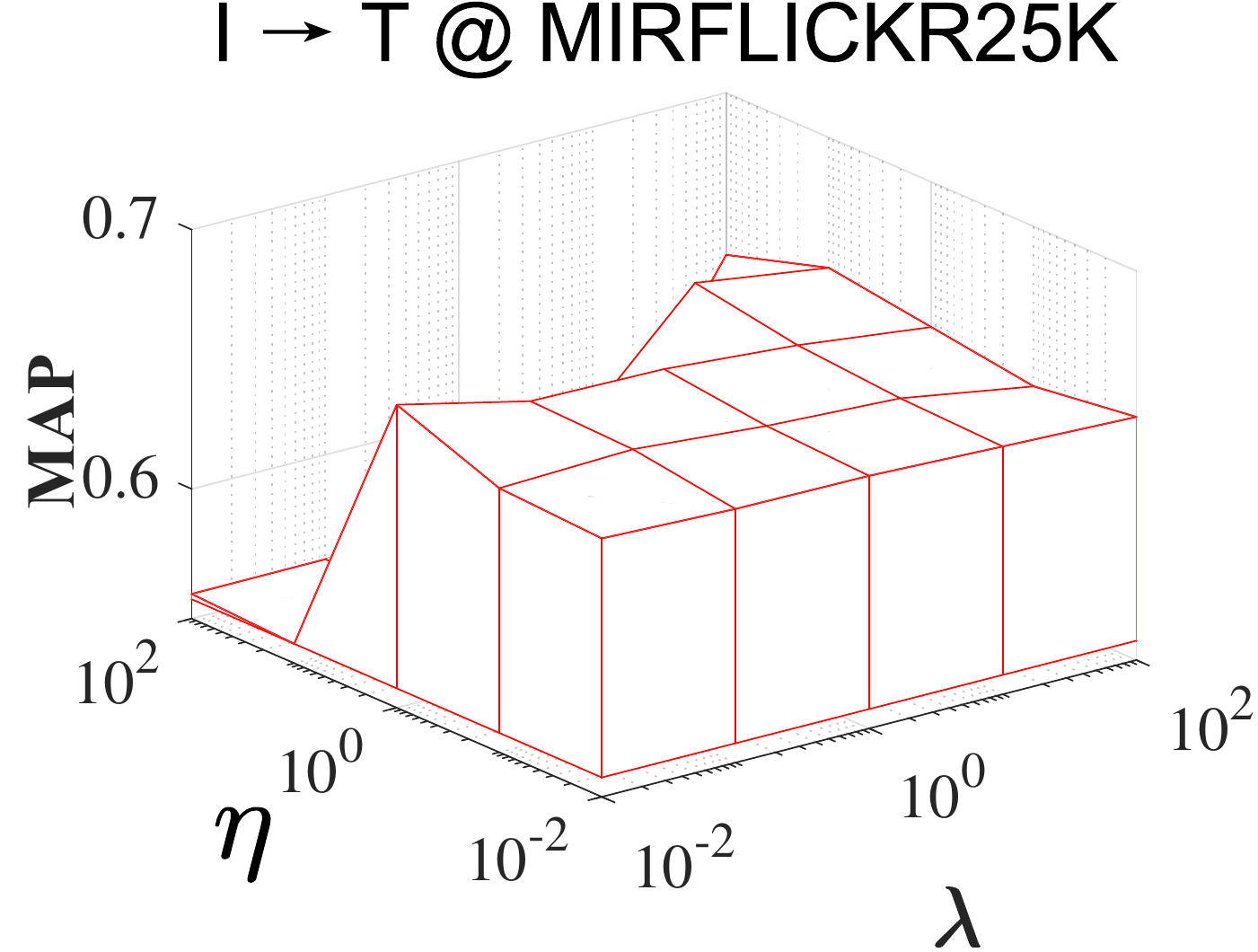}
\end{minipage}%
}%
\subfigure[]{
\begin{minipage}[t]{0.32\linewidth}
\centering
\includegraphics[width=1.55in]{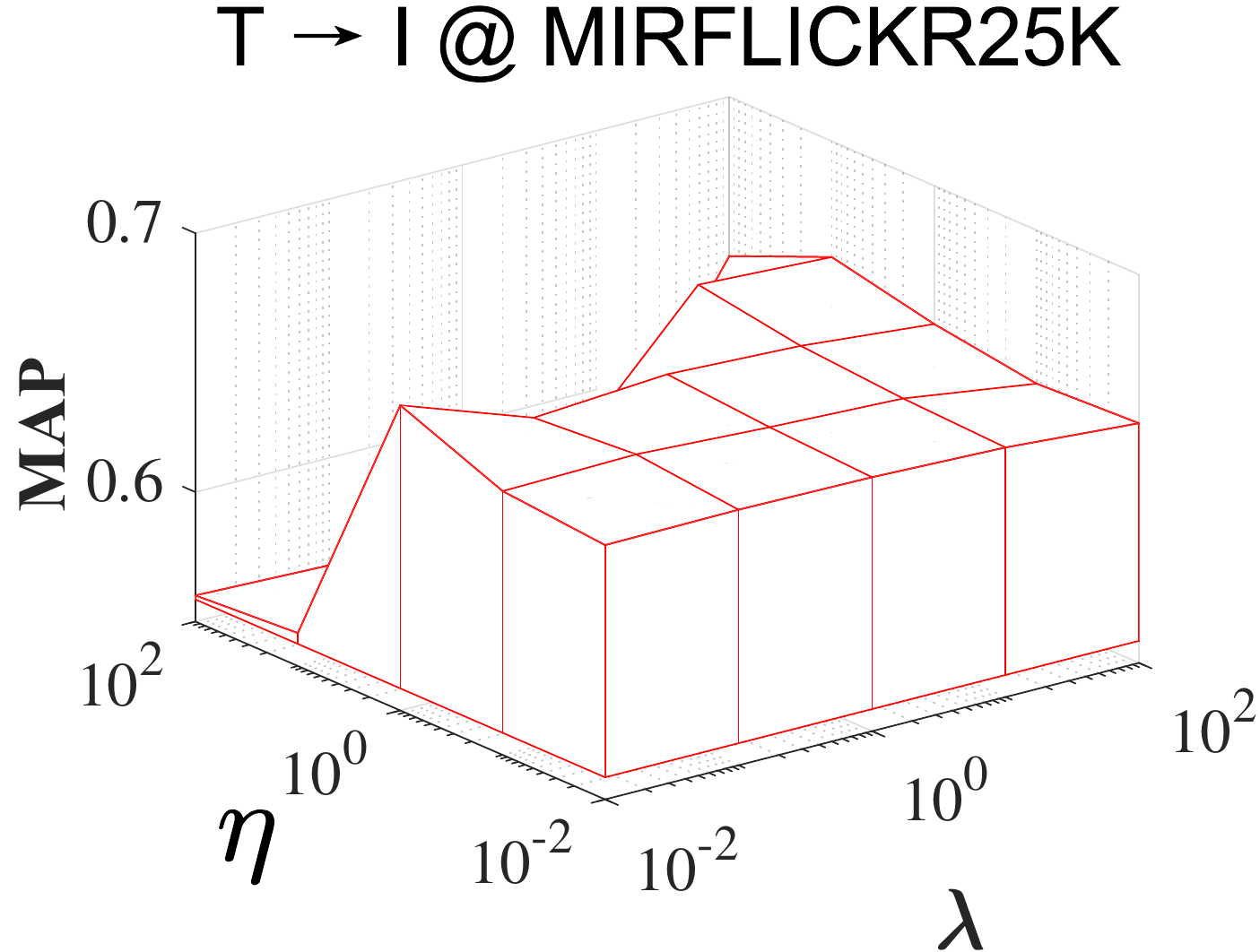}
\end{minipage}%
}%
\subfigure[]{
\begin{minipage}[t]{0.32\linewidth}
\centering
\includegraphics[width=1.55in]{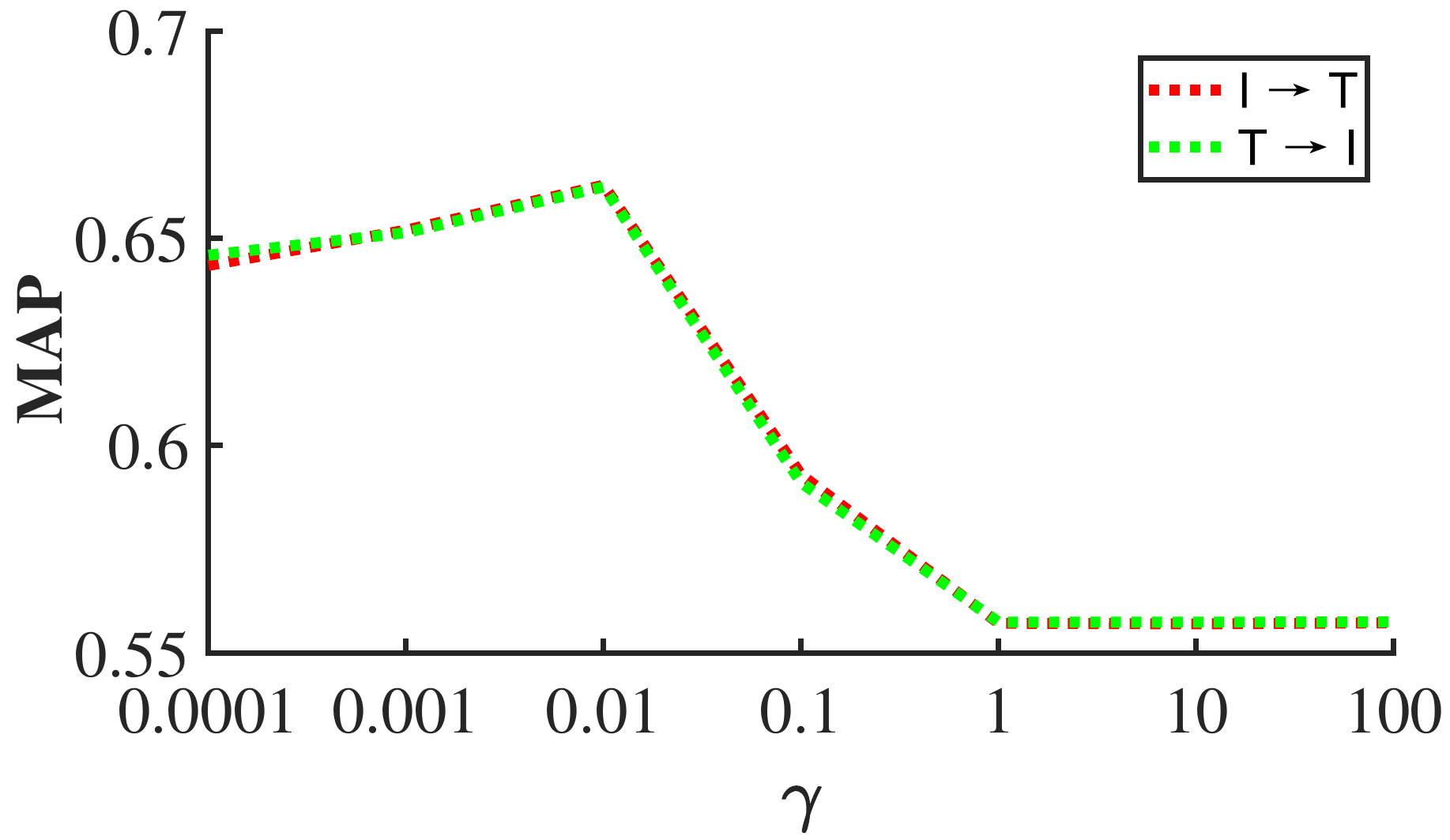}
\end{minipage}%
}%
\centering
\caption{Cross-modal retrieval accuracy of our DMH with different parameters setting in terms of MAP on MIRFLICKR-25K dataset: (a) and (b) show MAP with different $\lambda$ and $\eta$; whereas (c) shows MAP with different $\gamma$. The proportion of fully-paired data in train set is 90\%.}
\label{fig5}\end{figure}

In this part, we evaluate the parameter sensitivity of our DMH on MIRFLICKR-25K dataset at 16 bits with different hyper-parameter values. The proportion of fully-paired data in train set is 90\%. During the practice, we divide hyper-parameters into two groups according to their correlations: (1) $\lambda$;$\eta$, (2) $\gamma$. When we test the effect of one group of parameters, the other group is set to default values, i.e., $\lambda =\text{0.1}$, $\eta =\text{0.01}$ and $\gamma =0.01$ in Eq.~(13).

As shown in Figure~\ref{fig5}, the performance of our DMH is robust to hyper-parameters in a reasonable range. Specifically, Figure~\ref{fig5}~(a) and \ref{fig5}~(b) indicate that our DMH can achieve a satisfactory retrieval performance when $\lambda \in \left( 10^{-2}, 10^{2} \right)$ and $\eta \in \left( 10^{-2}, 10^0 \right)$. Meanwhile, Figure~\ref{fig5}~(c) also indicates that our method is robust to the variation of $\gamma$ within the range of $\left( 10^{-4}, 10^{-2} \right)$. However, when the values of $\eta$ and $\gamma$ are out of ranges, the performance of our method deteriorates dramatically. Therefore, cross validation is recommended to choose the appropriate hyper-parameters when using our method.

\begin{figure}[htbp]
\centering
\subfigure[Complemented Local Embedding~(CLE)]{
\begin{minipage}[t]{0.48\linewidth}
\centering
\includegraphics[width=1.55in]{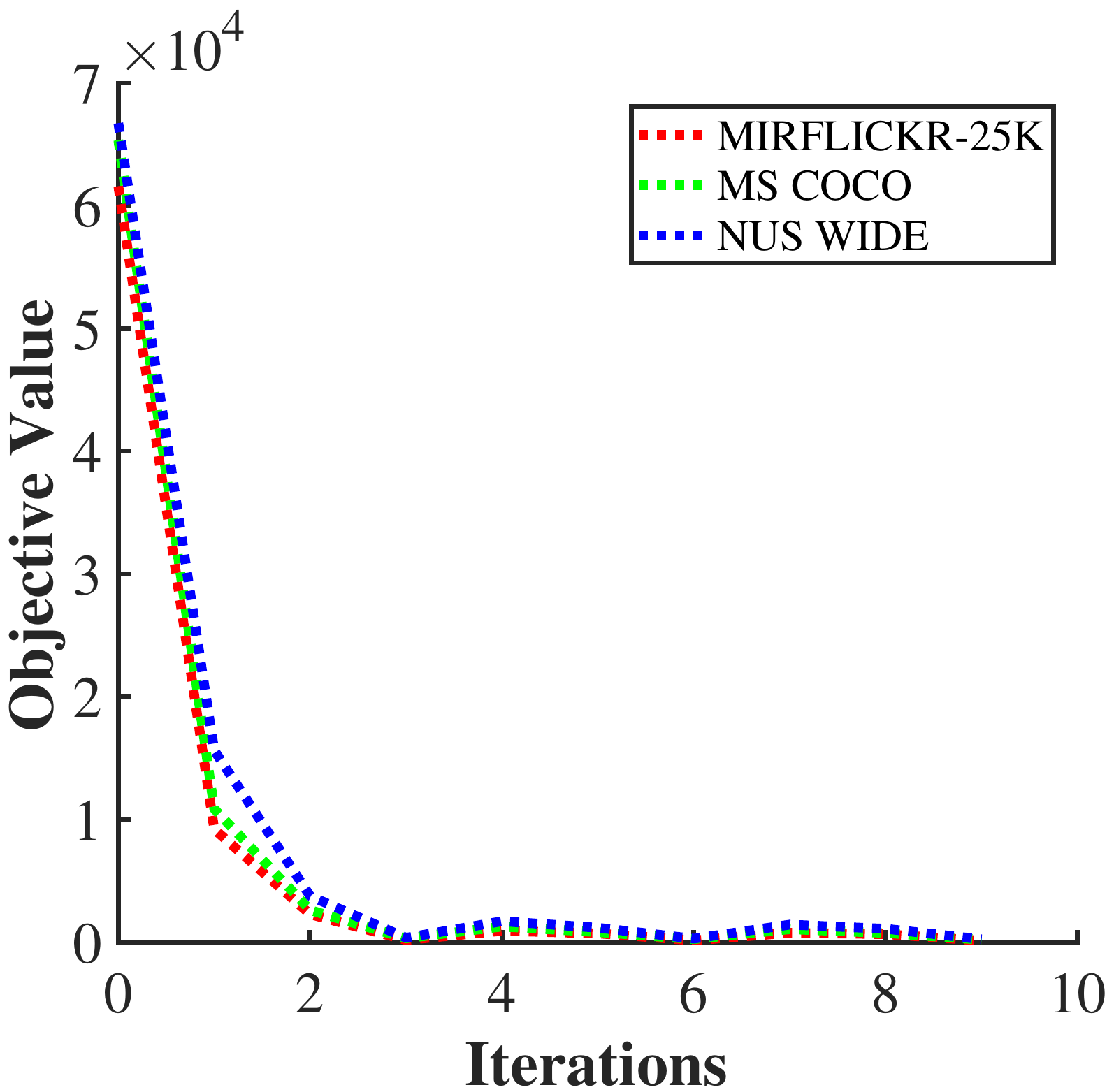}
\end{minipage}%
}%
\subfigure[Global Binary Embedding~(GBE)]{
\begin{minipage}[t]{0.48\linewidth}
\centering
\includegraphics[width=1.55in]{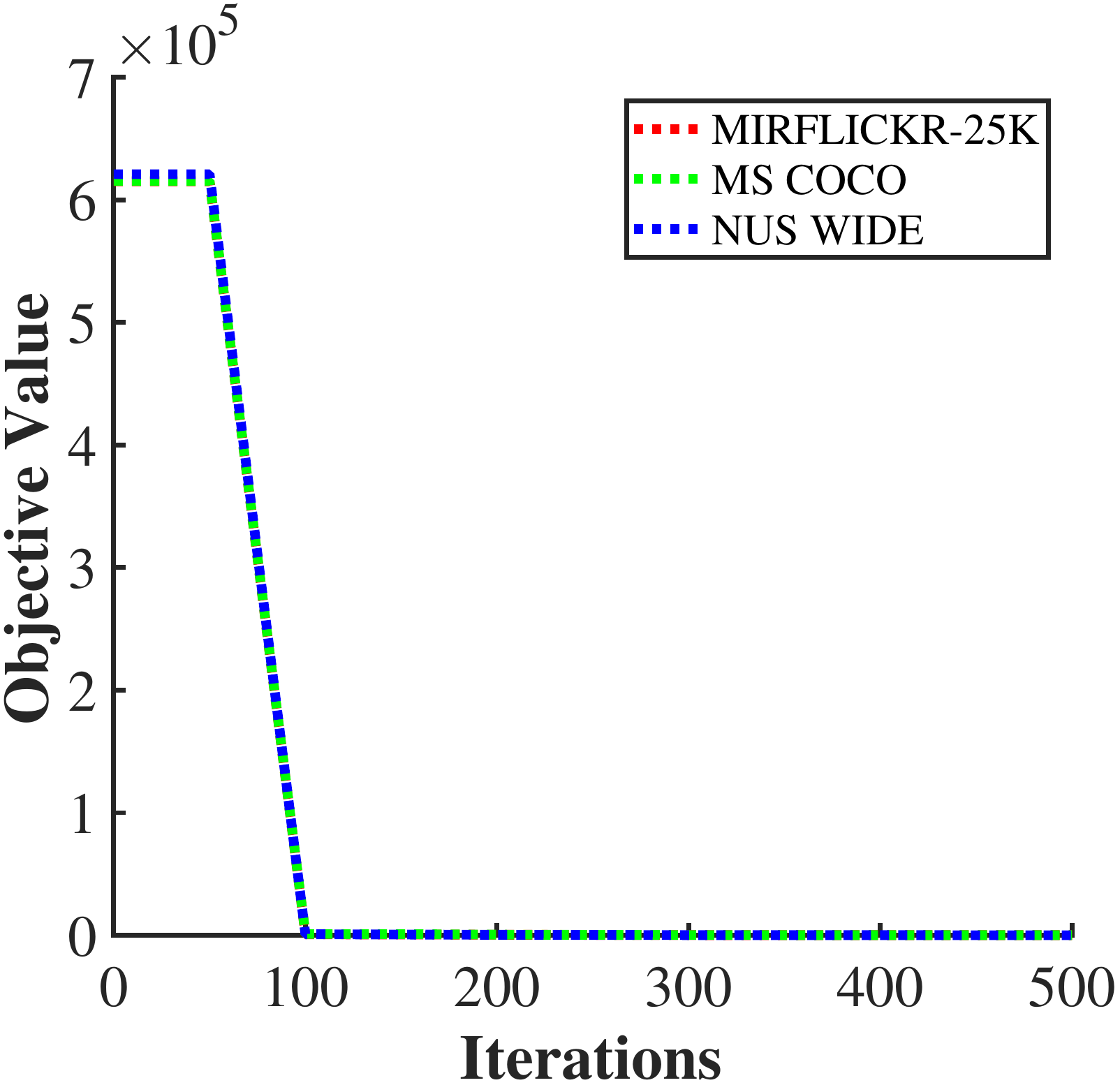}
\end{minipage}%
}%

\subfigure[Deep Adaptive Mapping~(DAM)]{
\begin{minipage}[t]{0.48\linewidth}
\centering
\includegraphics[width=1.55in]{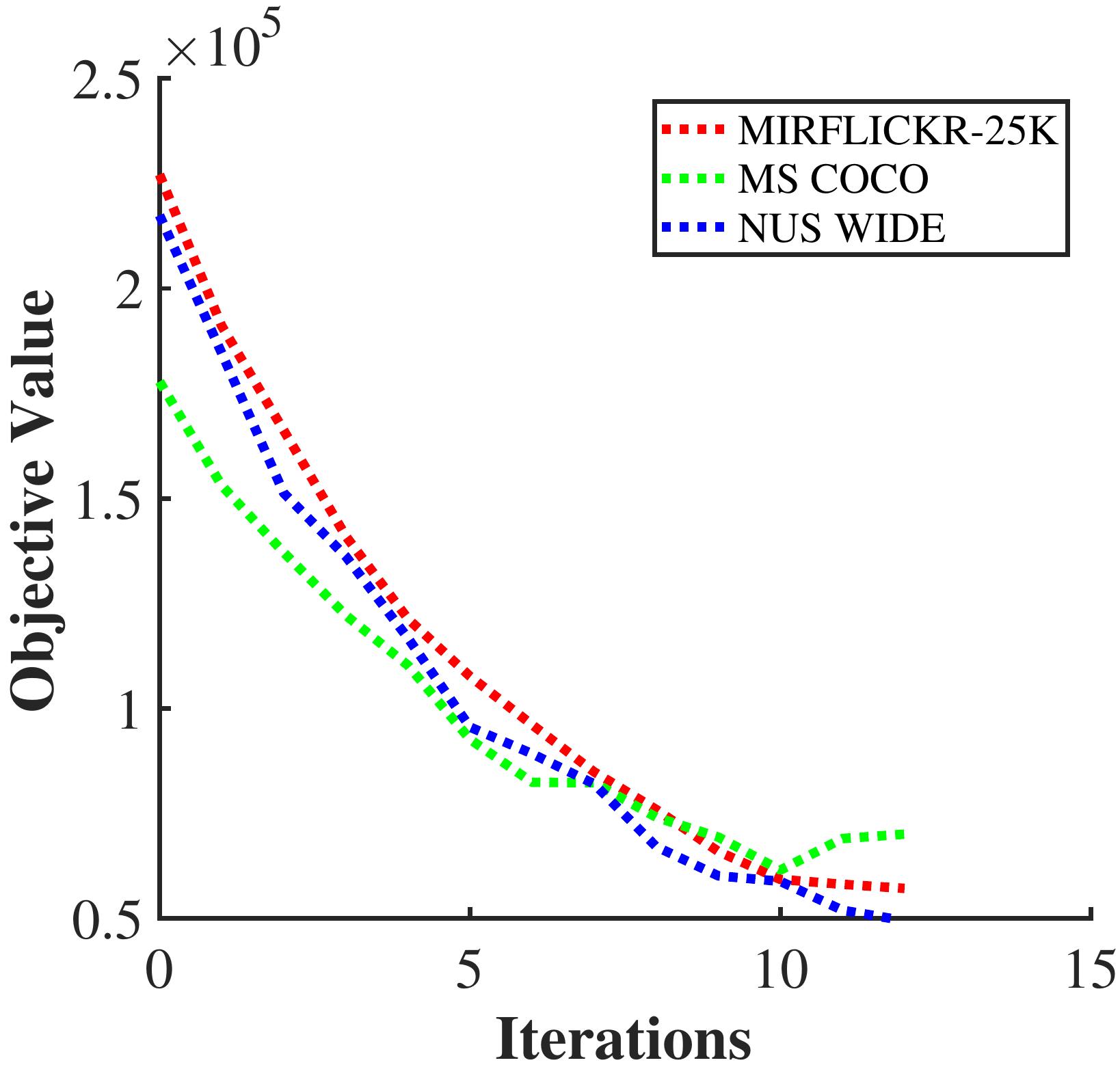}
\end{minipage}%
}%
\subfigure[Deep Manifold Hashing~(DMH)]{
\begin{minipage}[t]{0.48\linewidth}
\centering
\includegraphics[width=1.55in]{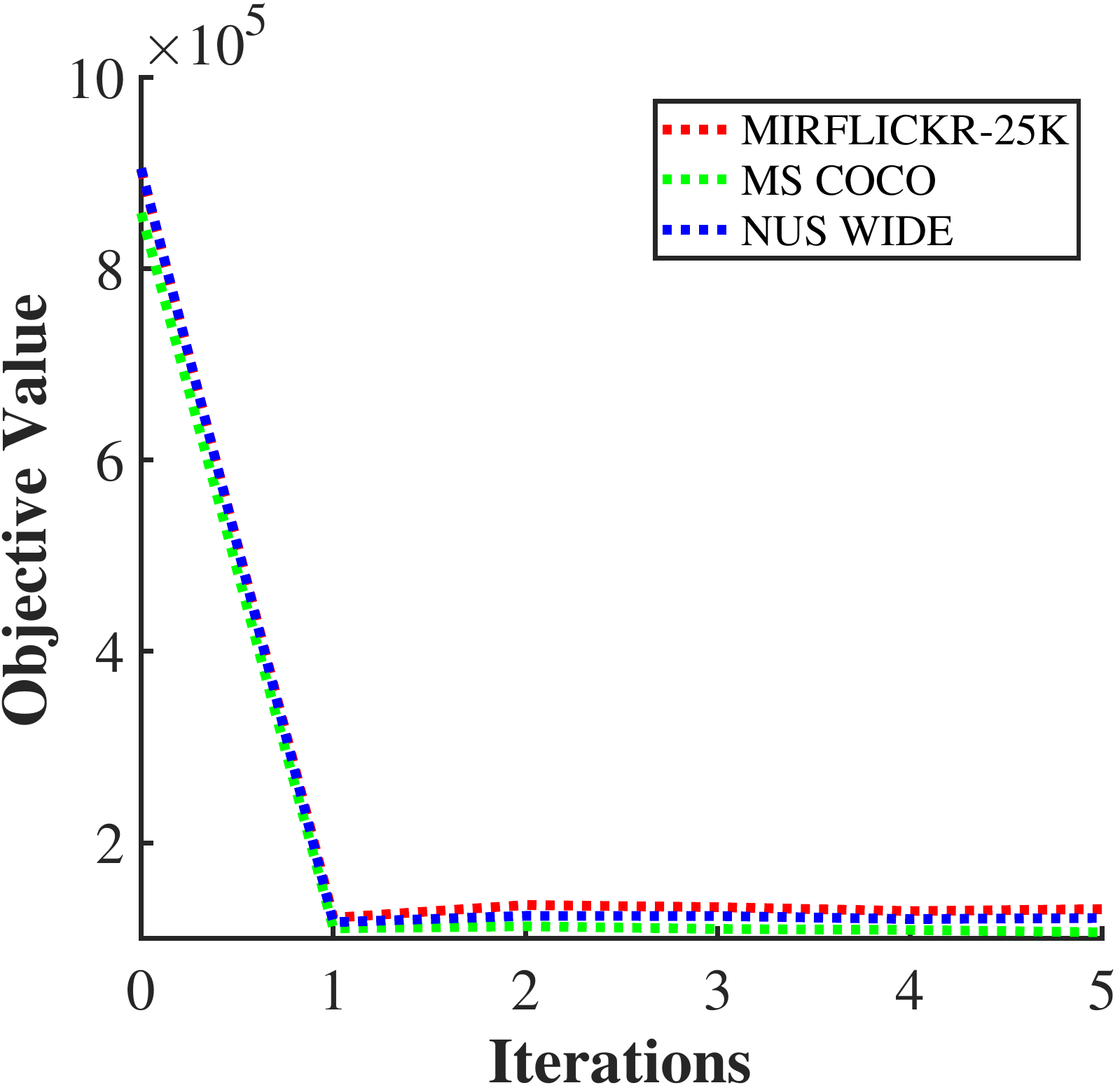}
\end{minipage}%
}%
\centering
\caption{Training error with the increase of iterations on three datasets at 16 bits: (a), (b) and (c) show the convergence curves of three sub-models in our DMH; whereas (d) shows the whole convergence curve of our method. The proportion of fully-paired data in train set is 90\%.}
\label{fig6}\end{figure}

Finally, convergence testing on MIRFLICKR-25K, MS COCO and NUS WIDE datasets at 16 bits are conducted to evaluate the convergence of our DMH. The proportion of fully-paired data in train set is 90\%. Since our method consists of three sub-models, we not only plot the convergence curve of overall model, but also report the testing results of every sub-model in Figure~\ref{fig6}. As can be seen, errors of three sub-models monotonically decrease with the increment of iterations in their corresponding training procedure, which leads the overall objective value of our DMH to be stable after 3 holistic iterations. It implies the fact that our DMH could converge within a limited number of iterations.
\section{Conclusion}
In this paper, inspired by the Divide-and-Conquer strategy, we present a novel Deep Manifold Hashing~(DMH) for semi-paired unsupervised cross-modal retrieval. Experimental results on three real-world datasets demonstrate that our DMH can more effectively remove modality gap compared with other state-of-the-arts, which thus improving retrieval accuracy. Moreover, our method also holds superior robustness in terms of different proportions of semi-paired data and the selection of parameters.

Despite the satisfactory results achieved by our DMH, more effects need to be taken on directly binary encoding. According to current results, our method relaxes the binary constraint in GBE to avoid the NP-hard optimization problem and thus leads to the sub-optimized hash codes. One feasible method to handle this is to explore matrix decomposition based discrete coding scheme instead of gradient descent based optimization methods. Some initial work has been started.

\section*{Acknowledgment}
This work was supported partially by the National Natural Science Foundation of China~(61571205 and 61772220), the Key Program for International S\&T Cooperation Projects of China~(2016YFE0121200), the Special Projects for Technology Innovation of Hubei Province~(2018ACA135), the Key Science and Technology Innovation Program of Hubei Province~(2017AAA017), the Natural Science Foundation of Hubei Province~(2018CFB691), fund from Science, Technology and Innovation Commission of Shenzhen Municipality~(JCYJ20180305180637611, JCYJ20180305180804836 and JSGG20180507182030600).

\bibliography{mybibfile}

\end{document}